\documentclass[twocolumn]{svjour3}         
\smartqed  
\usepackage{graphicx}

\usepackage{titlesec} 
\titleformat{\paragraph}[runin]
{\bfseries\scshape}{\theparagraph}{1em}{}
\titlespacing{\paragraph}{0em}{1ex}{.5em} 
\usepackage{todonotes}
\usepackage{booktabs}
\usepackage{tabularx}
\usepackage{multirow}
\usepackage{multicol}
\usepackage{bm}
\usepackage{url}
\usepackage{balance}
\usepackage{enumitem}
\usepackage{color}
\usepackage{bbding}
\usepackage{nccmath}
\usepackage{amsmath,amssymb}
\usepackage{graphicx}
\usepackage{cite}

\usepackage{graphicx}
\usepackage{amsmath}
\usepackage{amssymb}
\usepackage{color}
\usepackage{epstopdf}
\usepackage{array}
\usepackage{multirow}
\usepackage{amsfonts}
\usepackage{booktabs}
\usepackage{float}
\usepackage{subfigure}
\usepackage{algorithm}
\usepackage{algpseudocode}
\usepackage[switch]{lineno}
\usepackage[colorlinks,bookmarksopen,bookmarksnumbered,citecolor=blue,linkcolor=blue,urlcolor=black]{hyperref}

\begin{document}
\begin{sloppypar}
\title{Globally Correlation-Aware Hard Negative Generation}


\author{Wenjie Peng*\thanks{*~Both authors contributed equally to this research.} \and
        Hongxiang Huang* \and
        Tianshui Chen \and
        Quhui Ke \and \\
        Gang Dai \and
        Shuangping Huang\dag\thanks{\dag~Corresponding author.}
}


\institute{
          Wenjie Peng \at
             South China University of Technology, Guangzhou, China \\
              \email{eepwj@mail.scut.edu.cn}
           \and
          Hongxiang Huang \at
             South China University of Technology, Guangzhou, China \\
              \email{hhx769246124@gmail.com}
           \and
          Tianshui Chen \at
             Guangdong University of Technology, Guangzhou, China \\
              \email{tianshuichen@gmail.com}
           \and
          Quhui Ke \at
             South China University of Technology, Guangzhou, China \\
              \email{kquhui@gmail.com}
           \and
          Gang Dai \at
             South China University of Technology, Guangzhou, China \\
              \email{eedaigang@mail.scut.edu.cn}
           \and
          Shuangping Huang \at
             South China University of Technology, Guangzhou, China \\
              \email{eehsp@scut.edu.cn}
}

\date{Received: date / Accepted: date}

\maketitle


\def\eg{\emph{e.g.}}
\def\etal{\emph{et al}}

\newcommand{\EQREF}{Eq.~\eqref}
\newcommand{\EQSREF}{Eqs.~\eqref}
\newcommand{\FIGREF}{Fig.~\ref}
\def\proposed{VB} 
\def\fixcolor{black}
\def\hstate{\bm {\tilde s}}
\def\rstate{\bm s}
\def\jstate{\bm s^{jn}}
\def\jpolicy{\overrightarrow{\pi}}
\def\vpref{v_{\text {pref}}}
\def\vector#1{\mbox{\boldmath $#1$}}
\def\sup#1{^{(\rm #1)}}
\def\sub#1{_{\rm #1}}
\def\supi#1{^{(#1)}}
\def\vct#1{\mbox{\boldmath $#1$}}
\def\eg{{\it e.g.}}
\def\cf{{\it c.f.}}
\def\ie{{\it i.e.}}
\def\etal{{\it et al. }}
\def\etc{{\it etc}}
\newcommand{\argmax}{\mathop{\rm argmax}\limits}
\newcommand{\argmin}{\mathop{\rm argmin}\limits}

\def\Rerr{\Delta \bm r}
\def\Terr{\Delta \bm t}
\def\Xerr{\Delta \bm x}
\def\XerrRel{\Delta \bm {\tilde x}}
\def\Xgt{\dot{\bm x}}
\def\Rgt{\dot{R}}
\def\Tgt{\dot{\bm t}}
\def\arraystretchlen{1.0}

\def\cam{c}
\def\image{\mathcal I}
\def\traj{\mathcal X}
\def\btraj{\mathcal {\bm X}}
\def\keypoints{\mathcal P}
\def\states{\mathcal S}
\def\bstates{\mathcal {\bm S}}
\def\state{\bm s}
\def\ped{\bm x}
\def\pedi{\bm p} 
\def\obs{\bm z}

\def\Fi{\bm F_r}
\def\Fp{\bm F_p}
\def\vpref{\bm w}
\def\ENERGY{{\mathcal E}}

\def\DIFF#1{\textcolor{black}{#1}}
\def\DIFFCR#1{\textcolor{black}{#1}}

\begin{abstract}
Hard negative generation aims to generate informative negative samples that help to determine the decision boundaries and thus facilitate advancing deep metric learning. Current works select pair/triplet samples, learn their correlations, and fuse them to generate hard negatives. However, these works merely consider the local correlations of selected samples, ignoring global sample correlations that would provide more significant information to generate more informative negatives. In this work, we propose a Globally Correlation-Aware Hard Negative Generation (GCA-HNG) framework, which first learns sample correlations from a global perspective and exploits these correlations to guide generating hardness-adaptive and diverse negatives. Specifically, this approach begins by constructing a structured graph to model sample correlations, where each node represents a specific sample and each edge represents the correlations between corresponding samples. Then, we introduce an iterative graph message propagation to propagate the messages of node and edge through the whole graph and thus learn the sample correlations globally. Finally, with the guidance of the learned global correlations, we propose a channel-adaptive manner to combine an anchor and multiple negatives for HNG. Compared to current methods, GCA-HNG allows perceiving sample correlations with numerous negatives from a global and comprehensive perspective and generates the negatives with better hardness and diversity. Extensive experiment results demonstrate that the proposed GCA-HNG is superior to related methods on four image retrieval benchmark datasets.
\end{abstract}

\section{Introduction} 
Deep metric learning aims to design a metric for quantifying sample similarity in an embedding space, where similar samples cluster closely while the others go apart. This approach finds extensive application in various vision tasks, including image retrieval \cite{kim2023hier,Wang_2023_CVPR,li2020weakly,yu2021fine,husain2021actnet,liu2020learning}, face recognition \cite{schroff2015facenet,lu2017discriminative,yang2019learning}, and person re-identification \cite{li2022self,liao2022graph,zhu2023attribute}. One way to improve the performance of metric learning is to give more informative inputs for training. Hard negative generation (HNG) is one of the typical methods, which generates informative negative samples that help to determine the decision boundaries.

\begin{figure}
    \centering
    \includegraphics[width=\linewidth]{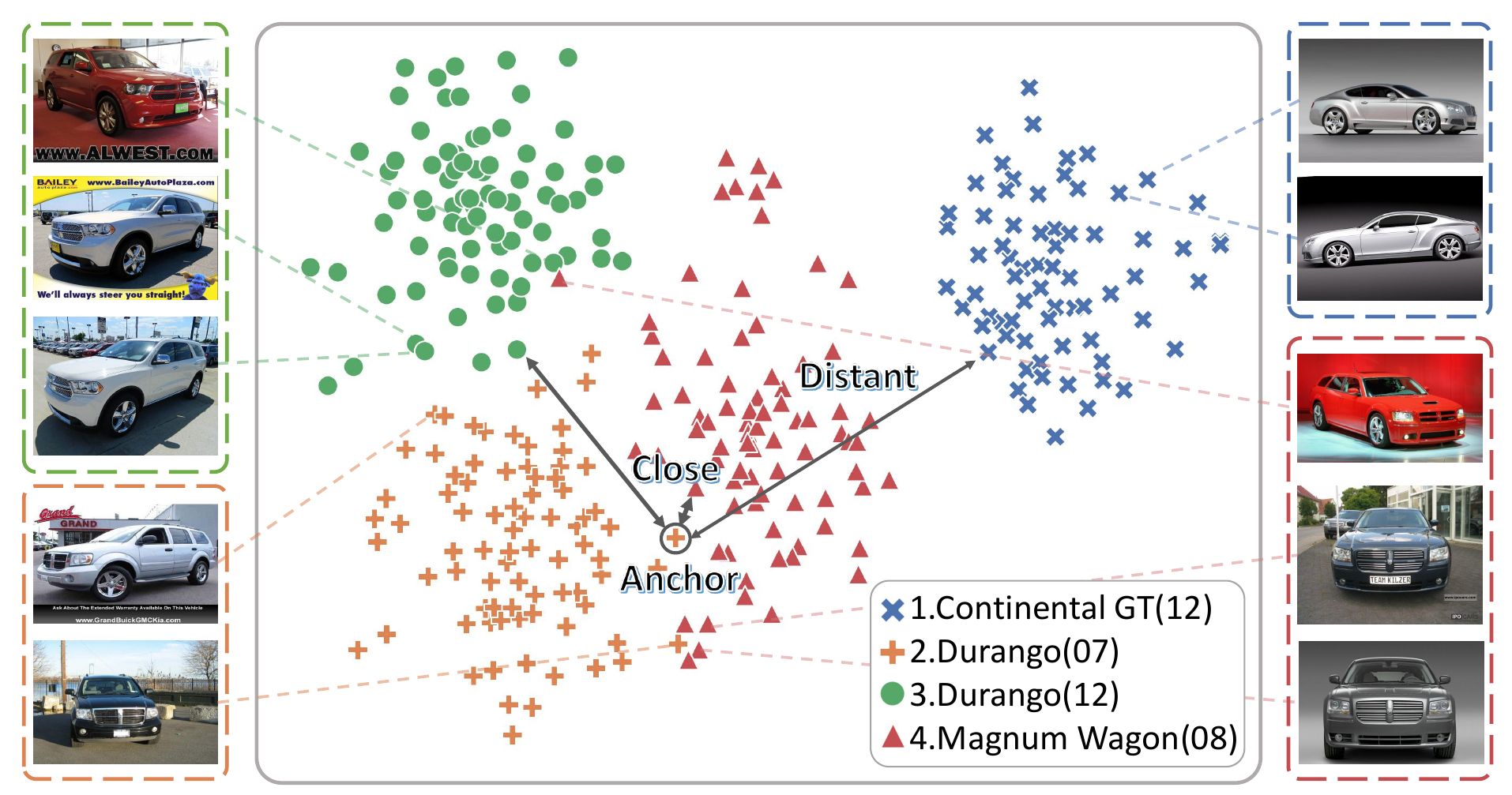}
    \caption{The t-SNE visualization depicts the embedding distributions of four classes on the Cars196 dataset, with each class represented by a distinct colored shape. This illustration emphasizes the importance of considering global correlations with negatives from diverse classes when analyzing a specific anchor for HNG. It points out the necessity of perceiving the global geometric distribution to generate harder synthetic negatives for closely related classes, thereby enhancing class discrimination. In contrast, for more distantly related negatives, it should control the hardness of the synthetic negative reasonably to avoid deviating from its corresponding class distribution.}
    \label{Fig_motivation}
\end{figure}

Current works on HNG typically fuse a pair or triplet of samples to produce a hard negative by GAN \cite{duan2019deep} or auto-encoder \cite{zheng2019hardness}. These methods primarily concentrate on local correlations within only a few selected samples, consequently leading to the synthesis of negatives with potentially inaccurate levels of hardness due to the lack of consideration for the potential influence of other classes. As illustrated in Fig. \ref{Fig_motivation}, consider an anchor belonging to class 2; selecting negative samples from closely related classes, such as 3 or 4, can effectively generate hard negatives to enhance the model discrimination between classes due to their relative proximity. In contrast, the anchor from class 2 and the samples from class 1 are significantly more distant. In scenarios where the HNG process only accounts for local sample correlations, there is a risk of generating synthetic negatives that inaccurately deviate from the intended embedding space of the corresponding negative class, potentially aligning with the distribution of an unrelated class, such as class 4. This misalignment underscores the critical necessity of incorporating the correlations of a boarder range of negative classes, \textit{i.e.}, global sample correlations, in the HNG process. By doing so, it becomes possible to accurately perceive the similarity among various negative classes, enabling the generation of informative negatives with appropriate levels of hardness and relevance.

In this work, we aim to explore sample correlations globally to help synthesize informative negative representations via a novel Globally Correlation-Aware Hard Negative Generation (GCA-HNG) framework. This framework perceives sample fine-grained correlations from a global perspective and subsequently harnesses these global correlations to guide informative negative generation. Although previous works \cite{zheng2019hardness,huang2020relationship,duan2019deep,zhu2022construct} have acknowledged the significance of HNG, they merely confine their focus to the local correlations within two or three samples and introduce a generator network to fuse them for synthesizing hard negatives via representation mapping. Instead, our proposed GCA-HNG framework extends beyond this narrow scope by learning global correlations across a broader range of negative classes, enabling it to accurately determine a sample's relative position within the global geometric distribution. Based on the learned correlations, GCA-HNG can synthesize negatives with adaptive hardness and rich diversity in a reasonable distribution range by a more flexible interpolation fusion approach. 

The GCA-HNG framework consists of two integral components: the Globally Correlation Learning (GCL) module and the Correlation-Aware Channel-Adaptive Interpolation (CACAI) module. Regarding the GCL module, we utilize a structured graph to model sample correlations within a mini-batch, where each node represents a specific sample and each edge represents the correlation between adjacent samples. To promote effective information interaction, we introduce an iterative graph message propagation mechanism. This mechanism facilitates a reciprocal exchange of information between nodes and edges, ensuring that every node and edge in the graph captures global context information and thus computes pairwise correlations globally. Moreover, we design the CACAI module to leverage the learned global correlation representations to produce channel-adaptive interpolation vectors for HNG. Unlike conventional methods \cite{zheng2019hardness,gu2020symmetrical,venkataramanan2022takes,ko2020embedding} that uniformly apply a single coefficient across all channels of the interpolation vector, our module determines each coefficient individually, based on the channel-level similarity between an anchor and a negative on a global scale. By leveraging these interpolation vectors, we perform interpolation between the designated anchor and multiple negatives from a specific class, yielding informative negative representations directly applicable in metric model optimization. This method eliminates the need for an additional generator network, enhancing both the efficacy and feature adaptability of the GCA-HNG framework. 

Our contributions are summarized as follows: 
\begin{itemize}
    \item We propose a novel GCA-HNG framework, which learns sample correlations globally and thus helps generate hardness-adaptive and diverse samples to facilitate deep metric learning. 
    \item We introduce a GCL module with perceptual associations with samples from more negative classes to determine the appropriate hardness for synthetic negatives, which enhances the alignment of synthetic representations with class labels. 
    \item We design a CACAI module to integrate global sample correlations to generate diverse negatives flexibly, improving the feature adaptability for HNG. 
    \item We conducted extensive experiments across four image retrieval benchmark datasets using various combinations of backbone networks and metric losses. The results show that GCA-HNG is superior to the current advanced methods. Codes and trained models are available at \url{https://github.com/PWenJay/GCA-HNG}. 
\end{itemize}

\section{Related Works}
\subsection{Deep Metric Learning}
Deep metric learning relies on backbone networks for feature extraction and employs various losses to learn the metrics. In this realm, the employment of backbones such as Convolutional Neural Networks (CNNs) \cite{he2016deep,ioffe2015batch,szegedy2015going} and Vision Transformers (ViTs) \cite{caron2021emerging,dosovitskiy2020image,touvron2021training} are pivotal for effective feature extraction. These backbones are instrumental in extracting meaningful features from input data. Metric losses, mainly including pairwise and proxy-based losses, are crucial for learning discriminative metrics, with each type of loss offering respective characteristics and variants. Pairwise losses, as elucidated by \cite{hadsell2006dimensionality,oh2016deep,wang2017deep,wang2019multi,weinberger2009distance,zhang2023denoising,dai2023disentangling,dai2025on,lin2024contrastive}, primarily focus on increasing the distance between negative pairs compared to positive pairs. The triplet loss \cite{weinberger2009distance} is a typical example of pairwise loss. It involves constructing a triplet consisting of an anchor, a positive, and a negative, with the objective of ensuring that the distance of the anchor-negative pair is greater than the distance of the anchor-positive pair by a predefined margin threshold $\alpha$. To surmount the limitation of only considering a single negative class in the triplet loss, Sohn et al. \cite{sohn2016improved} proposed the N-pair loss, which enables the tuple to interact with multiple negative classes for superior performance. Although effective in capturing intricate data-to-data correlations for similarity measurement, as visually supported by Zhu et al. \cite{zhu2021visual}, these methods still suffer from high training complexity. In contrast, proxy-based losses \cite{aziere2019ensemble,kim2020proxy,movshovitz2017no,qian2019softtriple,teh2020proxynca++,yang2022hierarchical} represent each class with a learnable proxy embedding, hence reducing the training complexity issues and considerably accelerating model convergence. The ProxyNCA loss \cite{movshovitz2017no} is the first approach to propose the proxy mechanism, adjusting loss constraints between an anchor sample and proxies. Subsequent iteration ProxyNCA++ \cite{teh2020proxynca++} further introduces several training strategies to improve model performance. Nevertheless, in the methods described above, each sample can only construct a distance measurement with the proxies and cannot take full advantage of the data-to-data correlations. To this end, Kim et al. \cite{kim2020proxy} proposed Proxy Anchor loss to represent the anchor with proxies rather than the positive or negative samples to harness data-to-data relationships more comprehensively. And it has achieved wide application in many current advanced works \cite{gu2021proxy,roth2022non,venkataramanan2022takes,zhang2022attributable,kim2023hier}. Moreover, some other types of metric losses are evolving simultaneously \cite{elezi2020group,elezi2022group,lim2022hypergraph}. As a result, there exist numerous combinations of backbones and metric losses used in deep metric learning, and we thoroughly evaluate the performance of GCA-HNG under various settings.

Recently, Graph Neural Networks (GNNs) \cite{kipfsemi,velivckovicgraph,kearnes2016molecular} have gained significant popularity due to their capacity to model and analyze intricate relationships. By representing data and their correlations through nodes and edges and exchanging information with their neighbors, GNNs can effectively capture contextual information from the entire graph and improve their representative ability. Researchers extensively deploy GNNs in many specific fields, including metric learning \cite{huang2020relationship,liao2022graph,seidenschwarz2021learning,zhang2022attributable,zhu2020fewer}, representing learning \cite{chen2022knowledge,chen2022cross,zeng2022keyword}, and image recognition \cite{chen2024heterogeneous,chen2024dynamic,chen2019multi,wang2020multi}. Numerous studies on metric learning employ graph structure to model data relationships and facilitate models to obtain better metrics. These relationships establish within a mini-batch \cite{seidenschwarz2021learning,zhang2022attributable}, across classes \cite{huang2020relationship,liao2022graph}, or between sample and proxies \cite{zhu2020fewer}. Inspired by the existing works in this field, this paper explores the integration of GNNs in modeling global sample correlations and assists the model in synthesizing more informative negatives. 

\subsection{Informative Input Construction}
In the domain of deep metric learning, the construction of informative inputs is crucial for distinguishing decision boundaries and acquiring discriminative metrics. Both mining-based and generation-based methods are the prevailing direction. Mining-based approaches, such as easy positive mining \cite{xuan2020improved}, hard negative mining \cite{bucher2016hard, schroff2015facenet, simo2015discriminative, gajic2021fast, zhu2019distance, suh2019stochastic} and hard example mining \cite{shrivastava2016training, smirnov2018hard, jin2018unsupervised}, have been proposed to oversample informative samples for efficient model convergence. Easy positive mining, as proposed by Xuan et al. \cite{xuan2020improved}, suggests that a query image need not be proximate to all the examples in its category but rather to a subset of easy positive examples within that category. This approach fosters a loosened learning strategy to obtain better generalization. In addition to easy positive mining, hard negative mining, explored in studies by \cite{bucher2016hard, schroff2015facenet, simo2015discriminative, gajic2021fast, suh2019stochastic}, aims to select the hard negative examples that exhibit high similarity to positive examples and frequently misclassified by the model during training. By emphasizing these challenging negatives in the training process, the model can enhance its ability to discriminate between positive and negative instances. Extending beyond these concepts, hard example mining, as proposed by \cite{shrivastava2016training, smirnov2018hard, jin2018unsupervised,tan2022cross,rao2023hierarchical}, is a broader concept that encompasses the selection of both challenging positive and negative examples. It aims to identify the most informative samples, regardless of their class, that can facilitate the acquisition of discriminative metrics. Despite their efficacy, these mining-based methods that oversampled a subset of the training set often suffer from limited sample diversity, potentially leading to overfitting and suboptimal generalization. 

Lately, recent advancements \cite{duan2019deep,zheng2019hardness,huang2022agtgan} have proposed the generation-based method to alleviate the problem of limited informative samples inherent in mining-based methods. Duan et al. \cite{duan2019deep} synthesized easy samples into hard negatives through adversarial training, thus increasing the utilization of easy samples. Zheng et al. \cite{zheng2019hardness} proposed hardness-aware representation interpolation between a given anchor and a negative, followed by an auto-encoder to generate the corresponding feature of hard negatives while preserving its class representation. This approach maximizes the utilization of information buried in all samples. Additionally, several studies \cite{zhao2018adversarial,zhu2022construct} explore hard triplet generation, which introduces a two-stage synthesis framework to generate hard positives and hard negatives simultaneously. However, these approaches focus on local correlations within a few samples and fail to consider global sample correlations. This oversight leads to potential misalignment of class labels with the embedding distributions of synthetic negatives, resulting in informational deficits.

In contrast to the above works, GCA-HNG introduces a novel approach utilizing a GNN based on the Transformer architecture to capture global sample correlations effectively. It achieves this via an innovative iterative graph message propagation mechanism, which assigns adaptive hardness to synthetic negatives based on the channel-level global correlations and further enhances their diversity. Notably, the representations of these synthetic negatives can be directly used for metric learning, circumventing the need for an additional generator network that involves feature alignment optimization \cite{zheng2019hardness,huang2020relationship,duan2019deep}. In this way, GCA-HNG demonstrates the ability to process features acquired from various backbone architectures, irrespective of the feature variance. We will detail this capability in Section \ref{experiment_bb}. The observation underscores the broader applicability of the GCA-HNG framework in deep metric learning.

\begin{figure*}
    \centering 
    \includegraphics[width=\linewidth]{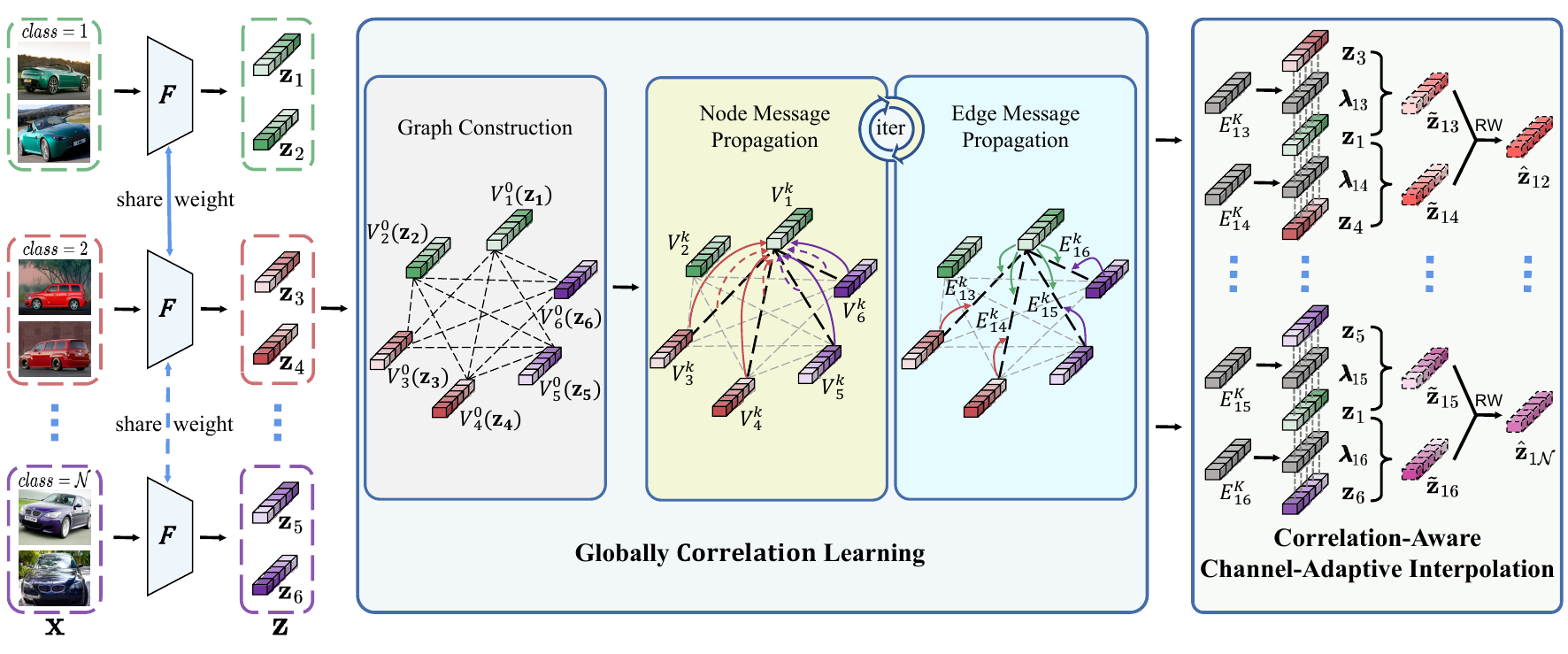}
    \caption{
    Schematic of the GCA-HNG framework. We use an embedding representation $\textbf{z}_1$ as the anchor and the embedding representations $\textbf{z}_3 \sim \textbf{z}_6$ as negatives to generate the hard negatives of $\textbf{z}_1$ as an illustrative example for clarity. GCA-HNG consists of a GCL module and a CACAI module. The former constructs a graph for representations $\textbf{z}$ and introduces an iterative graph message propagation mechanism, where node message propagation implements node-to-node and edge-to-node interaction to perceive global correlations, and edge message propagation implements node-to-edge interaction to facilitate the edges to model the correlations between adjacent samples globally. The nodes and edges are updated iteratively. The latter uses the learned global correlations $E_{1\cdot}^{K}$ to produce channel-adaptive interpolation vectors $\bm{\lambda}_{1\cdot}$ for each anchor-negative pair and integrates the anchor $\textbf{z}_1$ with multiple negatives from a specific class (\textit{e.g.}, $\textbf{z}_3,\textbf{z}_4$) by random weighting (RW) to generate the corresponding informative negatives (\textit{e.g.}, $\hat{\textbf{z}}_{12}$) of the anchor $\textbf{z}_1$. 
    }
    \label{fig_framework}
\end{figure*}

\section{Methodology}

\subsection{Overview}
\label{overview}
Let $\{(\textbf{x}_i,l_i)\}_{i=1}^{\mathcal{B}}$ represent a mini-batch of samples, formulated using a balanced sampling strategy \cite{zhai2018classification}. We organize each batch to incorporate $\mathcal{N}$ distinct classes, with each class comprising $m$ samples, thus establishing a batch size of $\mathcal{B}=\mathcal{N}\times m$. Each sample's label, $l_i$, belongs to the set of available classes $\{1,...,\mathcal{C}\}$, where $\mathcal{C}$ denotes the total number of classes. We arrange the samples into $m$ groups within the batch, with each group containing one sample from each of the $\mathcal{N}$ classes, maintaining a consistent order across all groups to ensure that the class labels $\{l_{i+g\mathcal{N}}\}_{g=0}^{m-1}$ stay consistent.

The GCA-HNG framework, depicted in Fig. \ref{fig_framework}, begins by transforming image data $\textbf{x}$ into embedding representations $\textbf{z}$ via a feature extractor $F$, which consists of a backbone network with a fully connected layer. Subsequently, it constructs a structured graph $\mathcal{G}$ from the embeddings $\textbf{z}$ to establish sample connectivity and introduces a GCL module to learn the correlations between samples globally. To achieve this, the GCL module utilizes a graph neural network $G$ equipped with an iterative graph message propagation mechanism, supported by the Transformer architecture, to promote the mutual exchange and aggregation of information between nodes and edges. This process allows each anchor to comprehensively consider its similarity with other negative samples within the global context of a mini-batch. Besides, the GCA-HNG framework incorporates a CACAI module, which converts the edge representations $E$ into adaptive interpolation vectors $\bm{\lambda}$. Each channel in these vectors corresponds to an adaptive interpolation coefficient, specifically designed to fuse channel-level information between an anchor and multiple negatives from a specific class. This module thus generates negative representations $\hat{\textbf{z}}$ that are both hardness-adaptive and diverse, enhancing the flexibility and effectiveness of the HNG process within metric learning.

\subsection{Globally Correlation Learning}

To effectively generate highly informative hard negatives, it is crucial to determine their appropriate hardness. This involves identifying a suitable interpolation point within each anchor-negative pair by assessing their similarity relative to other pairs in the global context of a mini-batch. To achieve this, we introduce a structured graph $\mathcal{G}$ to establish sample connectivity within a mini-batch, coupled with the employment of the graph network $G$ to enable learning sample correlations globally. 

Initially, we utilize a feature extractor $F$ to map a mini-batch of samples $\textbf{x}$ into embedding representations $\textbf{z}$. Using these embeddings, we construct a structured graph $\mathcal{G}=(V,E)$. In this graph, nodes $V$ represent the sample embeddings, and edges $E$ denote the correlations between corresponding samples. At the initialization stage, we specify the initial representations of each node and edge as $V_i^{k}|_{k=0}=\textbf{z}_i$, and $E_{i j}^{k}|_{k=0}=\textbf{z}_i \odot \textbf{z}_j$, respectively, with $k$ indicating the $k$-th iteration step.

\begin{figure*}
    \centering 
    \includegraphics[width=\linewidth]{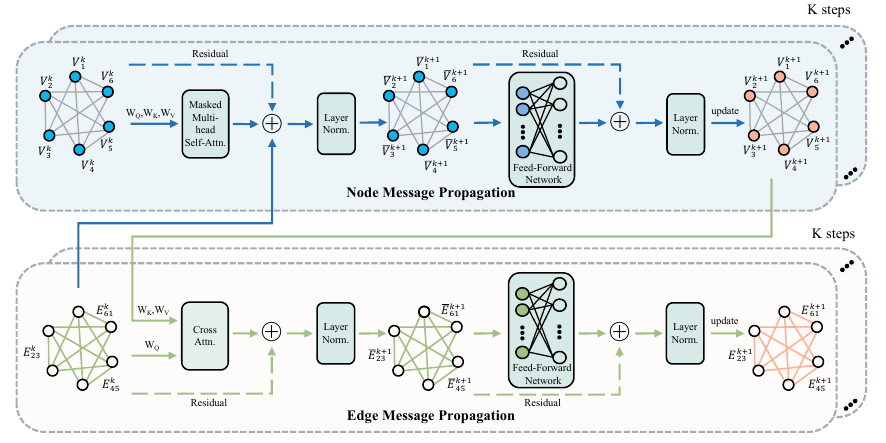}
    \caption{Implementation details diagram of the proposed iterative graph message propagation mechanism, where blue arrows indicate node message propagation routes and green arrows indicate edge message propagation routes.}
    \label{fig_propagation}
\end{figure*}

Subsequently, we introduce the graph network $G$, which learns sample correlations globally via a novel iterative graph message propagation. Detailed in Fig. \ref{fig_propagation}, this process sequentially updates node and edge representations within each iteration of the graph message propagation, starting with node message propagation and then followed by edge message propagation. This ordered updating ensures that node information effectively informs the subsequent edge updates, optimizing the network’s ability to capture and utilize global sample correlations comprehensively.

In terms of node message propagation, we incorporate a node-to-node interaction that allows the anchor to exclusively interact with all negative samples in the mini-batch while excluding interactions with itself and positive samples. This selective interaction utilizes a masked multi-head self-attention (MMSA) mechanism to act on the nodes, which is equivalent to sequentially treating each sample as an anchor that queries with the corresponding negatives acting as keys and values. Masking positive pairs is vital to prevent disproportionately high attention weights, which could reduce the model's ability to discern subtle differences between the anchor and various negative samples. Through this mechanism, each anchor gains a global perspective of anchor-negative similarities. Moreover, we augment the node representations by aggregating neighboring edge information into nodes through edge-to-node interactions, thus enriching the overall contextual understanding. Using the Transformer architecture \cite{vaswani2017attention}, we define the node message propagation mechanism as follows:
\begin{equation}
    \bar{V}_i^{{k+1}} \leftarrow \operatorname{\textit{LN}}\left(V_i^{k}+\operatorname{\textit{MMSA}}\left(V^{k}\right)_i+\sum_{j=1}^\mathcal{B} E_{i j}^{k}\right),
    \label{eq_node}
\end{equation}
\begin{equation}
    V_i^{k+1} \leftarrow \operatorname{\textit{LN}}\left(\operatorname{\textit{FFN}}\left(\bar{V}_i^{{k+1}}\right)+\bar{V}_i^{{k+1}}\right),
\end{equation}
where $\operatorname{\textit{MMSA}}\left(V^{k}\right)_i$ represents the output of the MMSA module for node $i$ during the $k$-th graph message propagation, $\operatorname{\textit{FFN}}$ refers to the feed-forward network, $\operatorname{\textit{LN}}$ denotes layer normalization.

Regarding edge message propagation, we introduce a node-to-edge interaction to fuse neighboring node representations (\textit{i.e.}, anchor, and negative) into edges through a cross-attention mechanism. In this setup, the edge representations serve as queries, while the node representations act as both keys and values. This process facilitates the incorporation of global information acquired in nodes into the edges, enabling the edges to capture the correlations of their adjacent nodes from a global perspective and further adjust attention tendencies between them. As a result, the edges can adaptively balance the hardness of the synthetic negatives between the anchor and negative. We also implement this edge message propagation mechanism using the Transformer architecture \cite{vaswani2017attention}, described as follows:
\begin{equation}
     \bar{E}_{i j}^{k+1} \leftarrow \operatorname{\textit{LN}}\left(E_{i j}^{k}+\operatorname{\textit{CA}}\left(E_{i j}^{k},V_i^{k+1},V_j^{k+1}\right)\right),
\end{equation}
\begin{equation}
     E_{i j}^{k+1} \leftarrow \operatorname{\textit{LN}}\left(\operatorname{\textit{FFN}}\left(\bar{E}_{i j}^{k+1}\right)+\bar{E}_{i j}^{k+1}\right),
\end{equation}
where $\operatorname{\textit{CA}}$ denotes the cross-attention mechanism.

After executing $K$ iterations of graph message propagation, the edges become thoroughly enriched with global sample correlations, significantly enhancing the informativeness of the resultant synthetic negatives. 

\subsection{Correlation-Aware Channel-Adaptive Interpolation}
Based on the above global sample correlations, we further design a channel-adaptive interpolation module that dynamically fuses the representations between an anchor and multiple negatives to synthesize informative negatives, which is essential for enhancing class discrimination. 

Firstly, we define the interpolation vector $\bm{\lambda}_{i j}$ for each anchor-negative pair ($\textbf{z}_i$, $\textbf{z}_j$) using their edge representation $E_{i j}^{K}$, expressed as follows:
\begin{equation}
     \bm{\lambda}_{i j}=\operatorname{\textit{Sigmoid}}\left(\textit{FC}(E_{i j}^{K})\right).
     \label{sigmoid}
\end{equation}
In this equation, \textit{FC} denotes a fully connected layer that transforms edge representation from the graph learning paradigm into interpolation coefficients, and the \textit{Sigmoid} function ensures these coefficients are normalized. The approach allows $\bm{\lambda}_{i j}$ to provide adaptive weight for channel-level embedding fusion within a specific anchor-negative pair. 

Secondly, our approach differs from the conventional interpolation method \cite{zheng2019hardness}, which uses a single interpolation coefficient based on local correlations between two samples. Instead, we propose a channel-adaptive interpolation mechanism, which incorporates global sample correlations to produce adaptive interpolation coefficients for different channels, thus capturing channel-level similarity correlations between the anchor and negative more effectively and expanding the search space of synthetic samples. We formulate this mechanism as follows:
\begin{align}
    \tilde{\textbf{z}}_{i j}= \begin{cases}\textbf{z}_i+\left[d^{+} + \bm{\lambda}_{i j}\eta (d^{-}-d^{+})
     \right] \frac{\textbf{z}_j-\textbf{z}_i}{d^{-}},&  \text {if } d^{-}>d^{+} \\ 
    \textbf{z}_j,& \text {if } d^{-} \leq d^{+}.\end{cases}
\label{eq_interpolation}
\end{align}
Here, $\textbf{z}_i$ denotes the anchor and $\textbf{z}_j$ the negative. $d^-$ and $d^+$ are the distances between the anchor-negative and anchor-positive pairs, respectively, calculated using normalized vectors. To progressively increase the hardness of synthetic negatives as the network learns, we introduce $\bm{\lambda}_{i j}\eta$, a dynamic scaling factor that modulates the interpolation point location. $\eta$ is set to $e^{-\frac{\alpha}{J_{ avg }}}$, where $J_{ avg }$, indicative of the current model's metric capabilities, is the average metric loss over the last epoch and $\alpha$ is a pulling factor \cite{zheng2019hardness}. As $J_{ avg }$ decreases, $\eta$ gradually tightens the upper limit of the interpolation interval, defined as $[0, \ \frac{\eta (d^{-}-d^{+})}{d^-}]$. Smaller values within this range indicate higher hardness for the interpolated negatives. The interpolation vector $\bm{\lambda}_{i j}$ produces suitable and deterministic values within this interval to achieve informative interpolation based on the learned global sample correlations. Using this method, we synthesize each anchor with all its respective negatives in a mini-batch successively to generate hard negatives. 

Furthermore, we diversify the synthetic negatives by fusing all interpolated representations between an anchor and the synthetic samples of a particular negative class using a random weighting approach:
\begin{equation}
     \hat{\textbf{z}}_{i n}=\operatorname{\textit{RW}}\left(\{\tilde{\textbf{z}}_{i j}|l_j=n\}\right),
\label{eq_rw}
\end{equation}
where $\operatorname{\textit{RW}}$ denotes the random weighting operation. Specifically, we fuse two elements within the set in Eq.(\ref{eq_rw}) by random weighting to obtain a new interpolated representation and then repeat the above operation between the interpolated one and the next element until all elements of the set are traversed. This method enables the model to explore a wider diversity of potentially hard negative $\hat{\textbf{z}}_{i n}$ between each anchor and negative class within a reasonable range. 

Notably, previous interpolation-based HNG methods \cite{zheng2019hardness,huang2020relationship,zhu2022construct} necessitate an additional generator module for representation mapping, converting the interpolated representations into feature space for metric learning. However, these methods encounter challenges when dealing with features that exhibit high variance, such as those from CNNs with max pooling or Vision Transformer architectures. In contrast, GCA-HNG directly uses interpolated representations for optimizing the metric model. This effectively circumvents the difficulties associated with optimizing the generator module for high-variance features, thereby broadening the applicability of GCA-HNG to various feature types. We will present a detailed proof in Section \ref{experiment_bb}.

\begin{algorithm*}
    \caption{GCA-HNG for Deep Metric Learning}
    \label{alg:algorithm}
    \textbf{Input}: $X$: training dataset. \\
    \textbf{Parameter}: feature extractor $F$, graph neural network $G$, classification heads $C_z$ and $C_v$. \\
    \textbf{Note}: In this algorithm, $sg(\cdot)$ denotes truncating gradients, thereby stopping the backpropagation to certain model parameters during specific steps.\\
    \textbf{Output}: The optimized parameters of feature extractor $F$, \textit{i.e.}, metric model. \\  
    \vspace{-\baselineskip}
    \begin{algorithmic}[1] 
        \While {not converged}
        \State Sample a mini-batch $\textbf{x}=\{(\textbf{x}_i,l_i)\}_{i=1}^{\mathcal{B}}$ from $X$.
        \State Map data $\textbf{x}$ to embedding space, \textit{i.e.}, $\textbf{z}=F(\textbf{x})$.
        \State Construct a graph $\mathcal{G}$ for embedding representations $\textbf{z}$ and predict the corresponding interpolation vectors based on Eq.(\ref{eq_node}-\ref{sigmoid}), \textit{i.e.}, $\bm{\lambda}=G(sg(\textbf{z}))$. 
        \State Employ $\bm{\lambda}$ to generate the representations of informative negative $\hat{\textbf{z}}$ based on Eq.(\ref{eq_interpolation}-\ref{eq_rw}).
        \State Update $G$ by the gradient on $J_{gen}$ based on Eq.(\ref{eq_Jgen}).
        \State Employ $sg(\textbf{z})$ to update $C_z$ by the gradient on $J_{cz}$ based on Eq.(\ref{eq_cz}).
        \State Construct a graph $\mathcal{G}$ for embedding representations $\textbf{z}$ and predict the corresponding interpolation vectors based on Eq.(\ref{eq_node}-\ref{sigmoid}), \textit{i.e.}, $\bm{\lambda}=G(\textbf{z})$.
        \State Employ $sg(\bm{\lambda})$ to generate the representations of informative negative $\hat{\textbf{z}}$ based on Eq.(\ref{eq_interpolation}-\ref{eq_rw}).
        \State Update $F$, $G$, and $C_v$ by the gradient on $J_m$ based on Eq.(\ref{eq_Jm}).
        \EndWhile
    \end{algorithmic}
\end{algorithm*}

\subsection{Learning Scheme}
We divide the optimization objective of the GCA-HNG framework into two distinct stages, which aim to guide the generation of hardness-adaptive and diverse negatives and seamlessly integrate them into various metric learning algorithms for end-to-end optimization. We have provided the detailed algorithmic process in Algorithm \ref{alg:algorithm}. 

\noindent
\textbf{Stage 1: Optimization of Hard Negative Generation}
\\
In the first stage of our GCA-HNG framework, we focus on guiding the generation of informative negatives using three distinct losses: classification loss, similarity loss, and diversity loss. Each targets a specific aspect of the generation process. These losses collectively enhance the model's ability to generate informative negatives, crucial for effective metric learning. 

Classification Loss $J_{ce}$: This loss employs a classification head $C_z$, optimized by real sample representations $\textbf{z}$, to ensure that generated negatives accurately preserve their respective class characteristics. We define the loss for each synthetic negative as follows: 
\begin{equation}
    J_{ce}=\operatorname{\textit{CE}}\left(C_z\left(\hat{\textbf{z}}_{i n}\right), l^{'}_n\right),
\end{equation}
where $\operatorname{\textit{CE}}$ represents the cross-entropy loss, and $l^{'}_n$ is the label for class $n$.

Similarity Loss $J_{sim}$: To enhance the challenge presented by synthetic negatives while maintaining their class relevance, we propose a similarity loss. This loss aims to increase the similarity between the anchor $\textbf{z}_i$ and the synthetic negative $\hat{\textbf{z}}_{in}$, calculated using the cosine similarity function:
\begin{equation}
    J_{sim}=1-\frac{\textbf{z}_i \cdot \hat{\textbf{z}}_{i n}}{\left\|\textbf{z}_i\right\|\left\|\hat{\textbf{z}}_{i n}\right\|}.
\end{equation}

Diversity Loss $J_{div}$: To broaden the diversity and expand the search space of synthetic negatives, we introduce a diversity loss based on the standard deviation of the interpolation vectors $\bm{\lambda}_{i \cdot}$ associated with each anchor $\textbf{z}_i$. This ensures a variety set of negatives for the model to learn from:
\begin{equation}
    J_{div}=1-\sigma\left(\bm{\lambda}_{i \cdot}\right),
\end{equation}
where $\sigma$ denotes the standard deviation function.

The overall objective for this stage is the simultaneous optimization of the graph network $G$ and the classification head $C_z$. To refine $G$, we focus on minimizing the combined objectives of these three losses, expressed as: 
\begin{equation}
    J_{gen}=\min _{G} \frac{1}{\mathcal{BN}} \sum_{i=1}^\mathcal{B} \sum_{\substack{n=l_1 ,\ n\neq l_i}}^{l_\mathcal{N}} (J_{ce}+\gamma_{s} J_{sim} + \gamma_{d} J_{div}),
\label{eq_Jgen}
\end{equation}
where $\gamma_{s}$ and $\gamma_{d}$ are weights for the similarity and diversity losses, respectively. This formulation illustrates the strategic supervision of interpolation vectors $\bm{\lambda}$ predicted by the network $G$ in class properties, hardness, and diversity. 

Concurrently, the optimization of the classification head $C_z$ focuses solely on the real sample representations $\textbf{z}$ to ensure accurate class prediction:
\begin{equation}
    J_{cz} = \min _{C_z} \sum_{i=1}^\mathcal{B} \operatorname{\textit{CE}}\left(C_z\left(\textbf{z}_i\right), l_i\right).
\label{eq_cz}
\end{equation}

\noindent
\textbf{Stage 2: Integration of Synthetic Negatives into Metric Learning}
\\
In the second stage of the GCA-HNG framework, the focus shifts to integrating synthetic hard negatives to strengthen discriminative training and achieve robust metric learning. This process also aids the graph network $G$ in learning robust node representations, which are crucial for generating high-quality hard negatives.

First, we incorporate $J_{gca}$, a cross-entropy loss that enables each node to retain its class semantics while understanding the global context within the mini-batch, expressed as:
\begin{equation}
    J_{gca}=\frac{1}{\mathcal{B}} \sum_{i=1}^\mathcal{B}\operatorname{\textit{CE}}\left(C_v\left(V_i^{K}\right), l_i\right),
\end{equation}
where $C_v$ is a classification head that processes the node outputs from the graph network $G$. 

Next, we introduce $J_{syn}$, a flexible pairwise loss that sequentially designates each sample within the mini-batch as an anchor, enabling interactions with corresponding positive and hard negative pairs. Unlike the N-pair loss, which merely assigns a specific group of samples as the anchor, $J_{syn}$ enhances the utilization of hard negatives by exploring richer correlations among them. We formulate this loss as follows:
\begin{equation}
    J_{syn}=\frac{1}{\mathcal{B}} \sum_{i=1}^\mathcal{B} \log \left(1+\sum_{\substack{n=l_1 ,\ n\neq l_i}}^{l_\mathcal{N}} \exp \left(\textbf{z}_i \hat{\textbf{z}}_{in}-\textbf{z}_i \textbf{z}^{+}_{i}\right)\right),
\end{equation}
where $\textbf{z}^{+}_{i}$ represents a positive representation selected from the same mini-batch. This strategy is crucial for facilitating the model to distinguish between closely matched positive and negative pairs. 

To verify the efficacy of the GCA-HNG framework, we utilize two types of metric losses: the N-pair (NP) loss \cite{sohn2016improved}, a classical pairwise loss, and the Proxy Anchor (PA) loss \cite{kim2020proxy}, a widely-used proxy-based loss. These losses, collectively termed $J_r$, are applied to real sample representations $\textbf{z}$. 

The original NP loss limits sampling to just two instances per class, designating the first group of samples as the anchor and the second as the positive pair. We have expanded this approach to allow more than two samples per class, thus enhancing sampling flexibility. This adjustment supports more extensive exploration of class perceptual scales and diversity effects for HNG. Our modification designates the first group of samples in a mini-batch as the anchor and subsequent $m-1$ groups sequentially as corresponding positive pairs. Notably, when sampling only two instances per class, our modified NP loss is equivalent to the original NP loss. We formulate this modified NP loss as follows:
\begin{align}
J_{r\_np}(\textbf{z}) =& \frac{1}{\mathcal{B}^{'}} \sum_{i=1}^{m-1} \sum_{j=1}^{\mathcal{N}} \log \Bigg(1+ \nonumber \\ 
& \sum_{q=1,\ q \neq j}^{\mathcal{N}} \exp \left(\textbf{z}_j \textbf{z}_{q+i\mathcal{N}}  -\textbf{z}_j \textbf{z}_{j+i\mathcal{N}}\right)\Bigg),
\end{align}
where $\mathcal{B}^{'}=(m-1)\mathcal{N}$. We have described the sample arrangement order in each mini-batch in Section \ref{overview}.

The formulation for the PA loss is as follows:
\begin{align}
J_{r\_pa}(\textbf{z})= & \frac{1}{\left|P^{+}\right|} \sum_{p \in P^{+}} \log \left(1+\sum_{\textbf{z}^{+} \in X_p^{+}} e^{-\alpha(s(\textbf{z}^{+}, p)-\delta)}\right) \nonumber \\ 
& +\frac{1}{|P|} \sum_{p \in P} \log \left(1+\sum_{\textbf{z}^{-} \in X_p^{-}} e^{\alpha(s(\textbf{z}^{-}, p)+\delta)}\right),
\end{align}
where $P$ denotes all proxy embeddings, with $P^{+}$ indicating all the positive proxies corresponding to the $\mathcal{N}$ sampling classes. $X^+_p$ and $X^-_p$ denote sets of representations that are, respectively, mutually positive and negative for the given proxy $p$. Parameters $\alpha$ and $\delta$ are scaling factor and margin, respectively, with $s(\cdot,\cdot)$ being the cosine similarity function.

The overall objective for this stage is to improve the discriminative capabilities of the graph network $G$ and the metric model $F$. We achieve this through a comprehensive strategy that integrates the above losses as follows:
\begin{equation}
    J_{m}=\min _{F, G, C_v} (J_{r}(\textbf{z})+J_{gca}+(1-\gamma_n) J_{syn}),
\label{eq_Jm}
\end{equation}
where $\gamma_n$ is set to $e^{-\frac{\beta}{J_{gen}}}$, serving as a balanced factor with $\beta$ as a hyper-parameter \cite{zheng2021hardness,zhu2022construct}. As network $G$ converges, it progressively increases the proportion of hard negatives to strengthen metric learning. 

\begin{table*}
\caption{Comparisons of image retrieval performance between our proposed GCA-HNG and current advanced deep metric learning methods on the CUB-200-2011 and Cars196 datasets. ``Arch." denotes the backbone architecture, ``Loss" where ``NP" indicates the original N-pair Loss and ``NP${}^{+}$" our modified version, and ``PA" for Proxy Anchor Loss. Bold indicates best. ${}^{\#}$: reproduced by us.}
\centering
\tabcolsep=1.3mm
\begin{tabular}{lcccccccccc}
\toprule
\multirow{2}{*}{Method} & \multirow{2}{*}{Arch.} & \multirow{2}{*}{Loss} & \multicolumn{4}{c}{CUB-200-2011} & \multicolumn{4}{c}{Cars196} \\ \cmidrule{4-11} 
& & & R@1 & R@2 & RP & \multicolumn{1}{c|}{M@R} & R@1 & R@2 & RP & M@R \\ \midrule
\midrule
DAMML \cite{duan2019deep} & G512 & NP & 53.9 & 66.7 & - & \multicolumn{1}{c|}{-} & 77.8  & 86.1 & - & - \\
HDML-A \cite{zheng2021hardness} & G512 & NP & 55.2 & 68.7 & - & \multicolumn{1}{c|}{-} & 81.1 & 88.8 & - & - \\
Symm \cite{gu2020symmetrical} & G512 & NP & 55.9 & 67.6 & - & \multicolumn{1}{c|}{-} & 76.5 & 84.3 & - & - \\
N-pair${}^{\#}$ \cite{sohn2016improved} & G512 & NP & 61.6$\pm$0.2 & 73.2$\pm$0.1 & 33.5$\pm$0.2 & \multicolumn{1}{c|}{22.4$\pm$0.1} & 78.1$\pm$0.2 & 86.5$\pm$0.1 & 34.5$\pm$0.1 & 23.6$\pm$0.1 \\
\textbf{GCA-HNG} & G512 & NP & 63.6$\pm$0.2 & 74.7$\pm$0.1 & 34.7$\pm$0.1 & \multicolumn{1}{c|}{23.6$\pm$0.1} & 83.5$\pm$0.1 & 90.1$\pm$0.1 & 37.5$\pm$0.2 & 27.1$\pm$0.2 \\
N-pair${}^{+}$ \cite{sohn2016improved} & G512 & NP${}^{+}$  & 62.0$\pm$0.3 & 73.2$\pm$0.1 & 33.5$\pm$0.3 & \multicolumn{1}{c|}{22.5$\pm$0.3} & 78.1$\pm$0.4 & 86.4$\pm$0.3 & 34.7$\pm$0.3 & 23.8$\pm$0.4 \\
\textbf{GCA-HNG} & G512 & NP${}^{+}$  & \textbf{63.9$\pm$0.1} & \textbf{75.2$\pm$0.1} & \textbf{34.9$\pm$0.2} & \multicolumn{1}{c|}{\textbf{23.8$\pm$0.2}} & \textbf{84.1$\pm$0.3} & \textbf{90.4$\pm$0.1} & \textbf{37.9$\pm$0.1} & \textbf{27.5$\pm$0.1} \\
\midrule  
N-pair${}^{+}$ \cite{sohn2016improved} \ & R512 & NP${}^{+}$  & 66.5$\pm$0.1 & 77.1$\pm$0.1 & 36.5$\pm$0.3  & \multicolumn{1}{c|}{25.5$\pm$0.3} & 84.8$\pm$0.1 & 90.7$\pm$0.2 & 39.5$\pm$0.1 & 29.3$\pm$0.1 \\
\textbf{GCA-HNG} & R512 & NP${}^{+}$ & \textbf{70.3$\pm$0.3} & \textbf{80.6$\pm$0.3} & \textbf{39.4$\pm$0.2} & \multicolumn{1}{c|}{\textbf{28.4$\pm$0.3}} & \textbf{88.2$\pm$0.3} & \textbf{93.1$\pm$0.2} & \textbf{42.3$\pm$0.4} & \textbf{32.5$\pm$0.4} \\
\midrule
N-pair${}^{+}$ \cite{sohn2016improved} \ & DN384 & NP${}^{+}$ & 76.7$\pm$0.2 & 85.2$\pm$0.2 & 43.6$\pm$0.2 & \multicolumn{1}{c|}{33.0$\pm$0.3} & 87.6$\pm$0.1 & 92.7$\pm$0.1 & 39.7$\pm$0.2 &  29.7$\pm$0.1\\
\textbf{GCA-HNG} & DN384 & NP${}^{+}$ & \textbf{78.5$\pm$0.1} & \textbf{86.4$\pm$0.1} &  \textbf{44.4$\pm$0.2} & \multicolumn{1}{c|}{\textbf{33.9$\pm$0.1}} & \textbf{90.0$\pm$0.1} & \textbf{94.4$\pm$0.1} & \textbf{43.6$\pm$0.1} & \textbf{34.1$\pm$0.1} \\
\midrule
Proxy Anchor${}^{\#}$ \cite{kim2020proxy} & G512 & PA & 64.7$\pm$0.1 & 75.2$\pm$0.1 & 35.0$\pm$0.1 & \multicolumn{1}{c|}{24.0$\pm$0.1} & 84.8$\pm$0.1 & 90.9$\pm$0.1 & 37.8$\pm$0.3 & 28.0$\pm$0.3 \\
\textbf{GCA-HNG} & G512 & PA & \textbf{65.7$\pm$0.1} & \textbf{76.8$\pm$0.2} & \textbf{35.8$\pm$0.1} & \multicolumn{1}{c|}{\textbf{24.8$\pm$0.1}} & \textbf{86.7$\pm$0.2} & \textbf{92.0$\pm$0.2} & \textbf{38.2$\pm$0.2} & \textbf{28.5$\pm$0.2}\\
\midrule 
Intra-Batch \cite{seidenschwarz2021learning} & R512 & PA & 70.3 & 80.3 & - & \multicolumn{1}{c|}{-} & 88.1 & 93.3 & - & - \\
Metrix \cite{venkataramanan2022takes} & R512 & PA & 71.0 & \textbf{81.8} & - & \multicolumn{1}{c|}{-} & 89.1 & 93.6 & - & - \\
NIR \cite{roth2022non} & R512 & PA & 70.5 & 80.6 & - & \multicolumn{1}{c|}{-} & 89.1 & 93.4 & - & - \\
HIER \cite{kim2023hier} & R512 & PA & 70.1 & 79.4 & - & \multicolumn{1}{c|}{-} & 88.2 & 93.0 &  - & - \\
HSE \cite{yang2023hse} & R512 & PA & 70.6 & 80.1 & - & \multicolumn{1}{c|}{-} & 89.6 & 93.8 &  - & - \\
Proxy Anchor${}^{\#}$ \cite{kim2020proxy}  & R512 & PA & 68.5$\pm$0.3 & 79.1$\pm$0.3 & 37.9$\pm$0.4 & \multicolumn{1}{c|}{27.2$\pm$0.4} & 88.4$\pm$0.1 & 93.4$\pm$0.2 & 40.3$\pm$0.2 & 30.9$\pm$0.2 \\
\textbf{GCA-HNG} & R512 & PA & \textbf{72.3$\pm$0.3} & 81.4$\pm$0.1  & \textbf{40.7$\pm$0.3} & \multicolumn{1}{c|}{\textbf{30.1$\pm$0.3}} & \textbf{91.1$\pm$0.2} & \textbf{94.8$\pm$0.1} & \textbf{44.1$\pm$0.3} & \textbf{35.4$\pm$0.3} \\
\midrule
HIER \cite{kim2023hier} & DN384 & PA & \textbf{81.1} & \textbf{88.2} & - & \multicolumn{1}{c|}{-} & 91.3 & 95.2 & - & - \\
Proxy Anchor${}^{\#}$ \cite{kim2020proxy}  & DN384 & PA & 79.0$\pm$0.3 & 86.9$\pm$0.2 & 45.0$\pm$0.2 & \multicolumn{1}{c|}{35.2$\pm$0.2} & 92.1$\pm$0.2 & 95.7$\pm$0.1 & 44.2$\pm$0.1 & 35.7$\pm$0.1 \\
\textbf{GCA-HNG} & DN384 & PA & 80.4$\pm$0.1 & 87.6$\pm$0.3 & \textbf{47.2$\pm$0.2} & \multicolumn{1}{c|}{\textbf{37.6$\pm$0.3}} & \textbf{93.1$\pm$0.1} & \textbf{96.3$\pm$0.1} & \textbf{44.8$\pm$0.2} & \textbf{36.5$\pm$0.2} \\
\bottomrule
\end{tabular}
\label{table_main1}
\end{table*}

\begin{table*}
\caption{Comparisons of image retrieval performance between our proposed GCA-HNG and current advanced deep metric learning methods on the SOP and InShop datasets. Given the sampling strategy of two samples per class for these datasets, our modified ``NP${}^{+}$" Loss becomes equivalent to the original ``NP'' Loss.}
\centering
\tabcolsep=1.0mm
\begin{tabular}{lcccccccccc}
\toprule
\multirow{2}{*}{Method} & \multirow{2}{*}{Arch.} & \multirow{2}{*}{Loss} &  \multicolumn{4}{c}{SOP} & \multicolumn{4}{c}{InShop} \\ \cmidrule{4-11} 
& & & R@1 & R@10 & RP & \multicolumn{1}{c|}{M@R} & R@1 & R@10 & RP & M@R \\ \midrule
\midrule
DAMML \cite{duan2019deep} & G512 & NP & 70.4 & 84.6 & - & \multicolumn{1}{c|}{-} & 80.8 & 94.6 & - & - \\
HDML-A \cite{zheng2021hardness} & G512 & NP & 70.7 & 85.0 & - & \multicolumn{1}{c|}{-} & 83.6 & 95.5 & - & - \\
Symm \cite{gu2020symmetrical} & G512 & NP & 73.2 & 86.7 & - & \multicolumn{1}{c|}{-} & - & - & - & -\\
N-pair${}^{\#}$ \cite{sohn2016improved} & G512 & NP/NP${}^{+}$ & 73.6$\pm$0.2 & 88.0$\pm$0.1 & 48.5$\pm$0.1 & \multicolumn{1}{c|}{45.0$\pm$0.1} & 86.1$\pm$0.1 & 97.0$\pm$0.1 & 62.5$\pm$0.2 & 59.3$\pm$0.2 \\
\textbf{GCA-HNG} & G512 & NP/NP${}^{+}$ & \textbf{75.8$\pm$0.1} & \textbf{89.6$\pm$0.1} &  \textbf{50.8$\pm$0.2} & \multicolumn{1}{c|}{\textbf{47.4$\pm$0.2}} & \textbf{88.6$\pm$0.1}& \textbf{97.7$\pm$0.1} & \textbf{66.6$\pm$0.2} & \textbf{63.5$\pm$0.2} \\
\midrule 
N-pair${}^{+}$ \cite{sohn2016improved} \ & R512 & NP${}^{+}$ & 78.1$\pm$0.1 & 90.4$\pm$0.1 & 53.8$\pm$0.2 & \multicolumn{1}{c|}{50.8$\pm$0.2} & 89.6$\pm$0.1 & 97.9$\pm$0.1 & 67.6$\pm$0.2 & 64.7$\pm$0.2 \\
\textbf{GCA-HNG} & R512 & NP${}^{+}$ & \textbf{80.5$\pm$0.1} & \textbf{91.9$\pm$0.1} & \textbf{56.8$\pm$0.1} & \multicolumn{1}{c|}{\textbf{53.9$\pm$0.1}} & \textbf{91.7$\pm$0.1} & \textbf{98.4$\pm$0.1} & \textbf{71.2$\pm$0.4} & \textbf{68.5$\pm$0.3} \\
\midrule
N-pair${}^{+}$ \cite{sohn2016improved} \ & DN384 & NP${}^{+}$ & 84.6$\pm$0.1 & 94.4$\pm$0.1 & 
63.5$\pm$0.2 & \multicolumn{1}{c|}{60.8$\pm$0.2} & 92.1$\pm$0.1 & 98.5$\pm$0.1 & 71.4$\pm$0.2 & 68.8$\pm$0.2 \\
\textbf{GCA-HNG} & DN384 & NP${}^{+}$ & \textbf{85.5$\pm$0.1} & \textbf{95.0$\pm$0.1} &  \textbf{65.2$\pm$0.1}
 & \multicolumn{1}{c|}{\textbf{62.6$\pm$0.1}} & \textbf{92.9$\pm$0.1} & \textbf{98.8$\pm$0.1} & \textbf{72.6$\pm$0.1} & \textbf{70.1$\pm$0.1} \\
\midrule
Proxy Anchor${}^{\#}$ \cite{kim2020proxy} & G512 & PA & 77.8$\pm$0.2 & 90.4$\pm$0.2 & 
53.3$\pm$0.3 & \multicolumn{1}{c|}{50.2$\pm$0.4} & 90.3$\pm$0.1 & 97.8$\pm$0.1 & 66.1$\pm$0.2 & 63.3$\pm$0.2 \\
\textbf{GCA-HNG} & G512 & PA &  \textbf{78.8$\pm$0.1} & \textbf{91.0$\pm$0.1} &  {\textbf{54.5$\pm$0.1}} & \multicolumn{1}{c|}{\textbf{51.4$\pm$0.1}} & \textbf{91.1$\pm$0.1} & \textbf{98.1$\pm$0.1} & \textbf{68.0$\pm$0.1} & \textbf{65.3$\pm$0.1} \\
\midrule 
Intra-Batch \cite{seidenschwarz2021learning} & R512 & PA & 81.4 & 91.3 & -
 & \multicolumn{1}{c|}{-} & 92.8 & \textbf{98.5} & - & - \\
Metrix \cite{venkataramanan2022takes} & R512 & PA & 81.3 & 91.7 & - & \multicolumn{1}{c|}{-} & 91.9 & 98.2& - & - \\
NIR \cite{roth2022non} & R512 & PA & 80.4 & 91.4 &  - & \multicolumn{1}{c|}{-} & - & - & - & - \\
HIER \cite{kim2023hier} & R512 & PA & 80.2 & 91.5 & - & \multicolumn{1}{c|}{-} & 92.4 & 98.2 &  - & - \\
HSE \cite{yang2023hse} & R512 & PA & 80.0 & 91.4 & - & \multicolumn{1}{c|}{-} & - & - &  - & - \\
Proxy Anchor${}^{\#}$ \cite{kim2020proxy}  & R512 & PA & 80.6$\pm$0.1 & 92.0$\pm$0.1 & 57.1$\pm$0.1 & \multicolumn{1}{c|}{54.2$\pm$0.1} & 92.2$\pm$0.1 & 98.3$\pm$0.1 & 69.8$\pm$0.1 & 67.2$\pm$0.1 \\
\textbf{GCA-HNG} & R512 & PA  & \textbf{81.8$\pm$0.1} & \textbf{92.4$\pm$0.1} &  \textbf{58.5$\pm$0.1} & \multicolumn{1}{c|}{\textbf{55.7$\pm$0.1}} & \textbf{93.6$\pm$0.1} & \textbf{98.5$\pm$0.1} & \textbf{71.9$\pm$0.2} & \textbf{69.5$\pm$0.2} \\
\midrule
HIER \cite{kim2023hier} & DN384 & PA & 85.7 & 94.6 & - & \multicolumn{1}{c|}{-} & 92.5 & 98.6  & - & - \\
Proxy Anchor${}^{\#}$ \cite{kim2020proxy}  & DN384 & PA & 85.2$\pm$0.2 & 94.4$\pm$0.2 & 
64.5$\pm$0.3 & \multicolumn{1}{c|}{62.0$\pm$0.2} & 92.0$\pm$0.2 & 98.3$\pm$0.1 & 67.0$\pm$0.3 & 64.4$\pm$0.3 \\
\textbf{GCA-HNG} & DN384 & PA & \textbf{85.9$\pm$0.1} & \textbf{94.8$\pm$0.1} & \textbf{ 65.6$\pm$0.1}  & \multicolumn{1}{c|}{\textbf{63.1$\pm$0.1}} & \textbf{93.3$\pm$0.1} & \textbf{98.7$\pm$0.1} & \textbf{70.8$\pm$0.2} & \textbf{68.3$\pm$0.2} \\
\bottomrule
\end{tabular}
\label{table_main2}
\end{table*}

\section{Experiments}
\subsection{Datasets and Evaluation Metrics}
\subsubsection{Datasets}
To thoroughly evaluate the GCA-HNG framework, we utilize four image retrieval benchmark datasets, including CUB-200-2011 \cite{welinder2010caltech}, Cars196 \cite{krause20133d}, Stanford Online Product (SOP) \cite{oh2016deep}, and InShop Clothes (InShop) \cite{liu2016deepfashion}. We follow the settings for dataset partitioning as delineated in \cite{kim2020proxy,seidenschwarz2021learning,zheng2021hardness}. Below are detailed descriptions of these datasets:
\begin{itemize}
    \item \textbf{CUB-200-2011 Dataset.} Proposed by Welinder et al. \cite{welinder2010caltech}, this dataset contains 11,788 images across 200 bird species. The training set comprises the first 100 species with 5,864 images, and the testing set includes the rest 100 species with 5,924 images.
    \item \textbf{Cars196 Dataset.} Introduced by Krause et al. \cite{krause20133d}, this dataset comprises 16,185 images representing 196 car models. The division entails a training set of 98 models with 8,054 images, and a testing set of 98 models incorporating 8,131 images.
    \item \textbf{Stanford Online Products Dataset.} Presented by Oh Song et al. \cite{oh2016deep}, this dataset consists of 120,053 images of 22,634 products from eBay.com. The training set includes the first 11,318 products with 59,551 images, while the testing set contains the rest 11,316 products with 60,502 images.
    \item \textbf{InShop Clothes Dataset.} Developed by Liu et al. \cite{liu2016deepfashion}, this dataset comprises 52,712 images of 7,982 clothes items from an online shopping website. The training set consists of 3,997 categories with 25,882 images. The testing set includes a query subset with 14,218 images of 3,985 clothing items and a gallery subset with 12,612 images of 3,985 items.
\end{itemize}

The four benchmark datasets include only training and testing sets, lacking a validation set. Although Musgrave et al. \cite{musgrave2020metric} proposed a standard evaluation protocol for partitioning training data to create a validation set, most top-tier studies rarely adopt this approach. The reasons include the scarcity of studies reporting results under this protocol, hindering direct performance comparisons, the reduction in training data often leading to inferior performance, and the lack of InShop results. To address this, we follow the common practice in metric learning of treating the testing set as the validation set. While not the optimal choice, it ensures fair comparisons with studies adopting the same protocol.

\subsubsection{Evaluation Metrics}
Following the setting proposed by Musgrave et al. \cite{musgrave2020metric}, we evaluate the retrieval performance of the metric model using three metrics: Recall@Ks (R@K), R-Precision (RP), and Mean Average Precision at R (M@R). R@K measures the probability that the correct class appears within the Top-K retrieved results for each query. RP evaluates the precision at the rank R, where R is the total number of references sharing the same class as the query. M@R calculates the average of the precision values obtained after each relevant result is retrieved up to the R-th rank. These metrics collectively offer a comprehensive evaluation of the model's retrieval capacity in precision and recall.

\subsection{Implementation Details}
In our study, we validate the GCA-HNG framework using prevalent CNN backbones, including GoogLeNet \cite{szegedy2015going} and ResNet-50 \cite{he2016deep}, which are widely utilized in the metric learning field \cite{zheng2019hardness,gu2020symmetrical,venkataramanan2022takes,roth2022non}. Following recent advanced studies \cite{ermolov2022hyperbolic,kim2023hier}, we also incorporate a ViT backbone, \textit{i.e.}, DINO-Small \cite{caron2021emerging}, to demonstrate the framework's broad applicability. All these backbones are pre-trained on the ImageNet ILSVRC dataset \cite{russakovsky2015imagenet}. For CNNs, we apply specific pooling operations to reduce spatial dimensions to a single vector: ResNet-50 utilizes adaptive average pooling combined with max pooling, while GoogLeNet employs average pooling. For ViTs, we use the CLS token's embedding as the output. Subsequently, to ensure a fair comparison with current research \cite{zheng2021hardness,roth2022non,kim2023hier}, we add a fully connected layer to each backbone, mapping the output dimensions to 512, 512, and 384 for GoogLeNet, ResNet-50, and DINO, respectively, denoted as G512, R512, and DN384 in Tables \ref{table_main1} and \ref{table_main2}. 

We ensure uniform image preprocessing across all experiments: during training, we randomly crop and resize input images to 224$\times$224 and apply a random horizontal flipping augmentation. For testing, we resize images to 256$\times$256 and center crop them to 224$\times$224. We construct mini-batches using a balanced sampling strategy with a batch size of approximately 80, processed on an NVIDIA RTX 3090 GPU. Specifically, for CUB-200-2011 and Cars196 datasets, we employ a 27$\times$3 sampling strategy detailed in Section \ref{class_sampling_sec}. For SOP and InShop datasets, we adopt a 40$\times$2 strategy due to many classes having fewer than three samples.

We optimize the network using the AdamW optimizer \cite{loshchilov2018decoupled} with a weight decay of $10^{-4}$. Hyper-parameter settings include: $\alpha=5$, $\beta=2$, $\gamma_d=0.01$ for PA loss and $0.03$ for NP loss, and $\gamma_s=1$. The graph network $G$ applies a cosine decay strategy during optimization. Learning rates for the feature extractor $F$, graph network $G$, and classification heads $C_z$ and $C_v$ are set to $1.5\cdot10^{-4}$, $3\cdot10^{-4}$, $10^{-3}$, and $3\cdot10^{-4}$, respectively. We unify the graph message propagation steps $K$ and the number of heads $H$ in the MMSA module to 2 and 4 for the CUB-200-2011 and Cars196 datasets, and 1 and 2 for the SOP and InShop datasets, respectively. Section \ref{KH_exploration_sec} provides the rationale for these choices. 

\subsection{Quantitative Results}
Based on our implementation settings, we provide quantitative results from deploying the GCA-HNG framework with PA and our modified NP losses, using GoogLeNet, ResNet-50, and DINO as backbones. As shown in Tables \ref{table_main1} and \ref{table_main2}, we present the performance of GCA-HNG across the CUB-200-2011, Cars196, SOP, and InShop datasets. We compare GCA-HNG with current advanced deep metric learning methods, including DAMML \cite{duan2019deep}, HDML-A \cite{zheng2021hardness}, Symm \cite{gu2020symmetrical}, Intra-Batch \cite{seidenschwarz2021learning}, Metrix \cite{venkataramanan2022takes}, NIR \cite{roth2022non}, HSE \cite{yang2023hse}, and HIER \cite{kim2023hier}. In particular, for a fair comparison with prior studies under the G512 architecture, we validated our framework using both the original and our modified NP losses. The results demonstrate that GCA-HNG comprehensively exceeds all baseline configurations and outperforms most current leading competitors across these benchmarks. For instance, using ResNet-50 with the modified N-pair loss, GCA-HNG achieves $70.3\%$ in Recall@1, $39.4\%$ in RP, and $28.4\%$ in M@R on the CUB-200-2011 dataset, marking improvements of $3.8\%$ in Recall@1 and $2.9\%$ in both RP and M@R compared to the baseline. Further experiments across other configurations substantiate the robust performance of GCA-HNG in diverse scenarios. This success primarily stems from its global correlation perception with various negative classes for synthesizing informative negatives, thus boosting the metric model's retrieval performance and confirming its efficacy in metric learning.

\subsection{Qualitative Results}
\begin{figure*}
    \centering
    \subfigure[CUB-200-2011]{
        \includegraphics[width=0.48\linewidth]{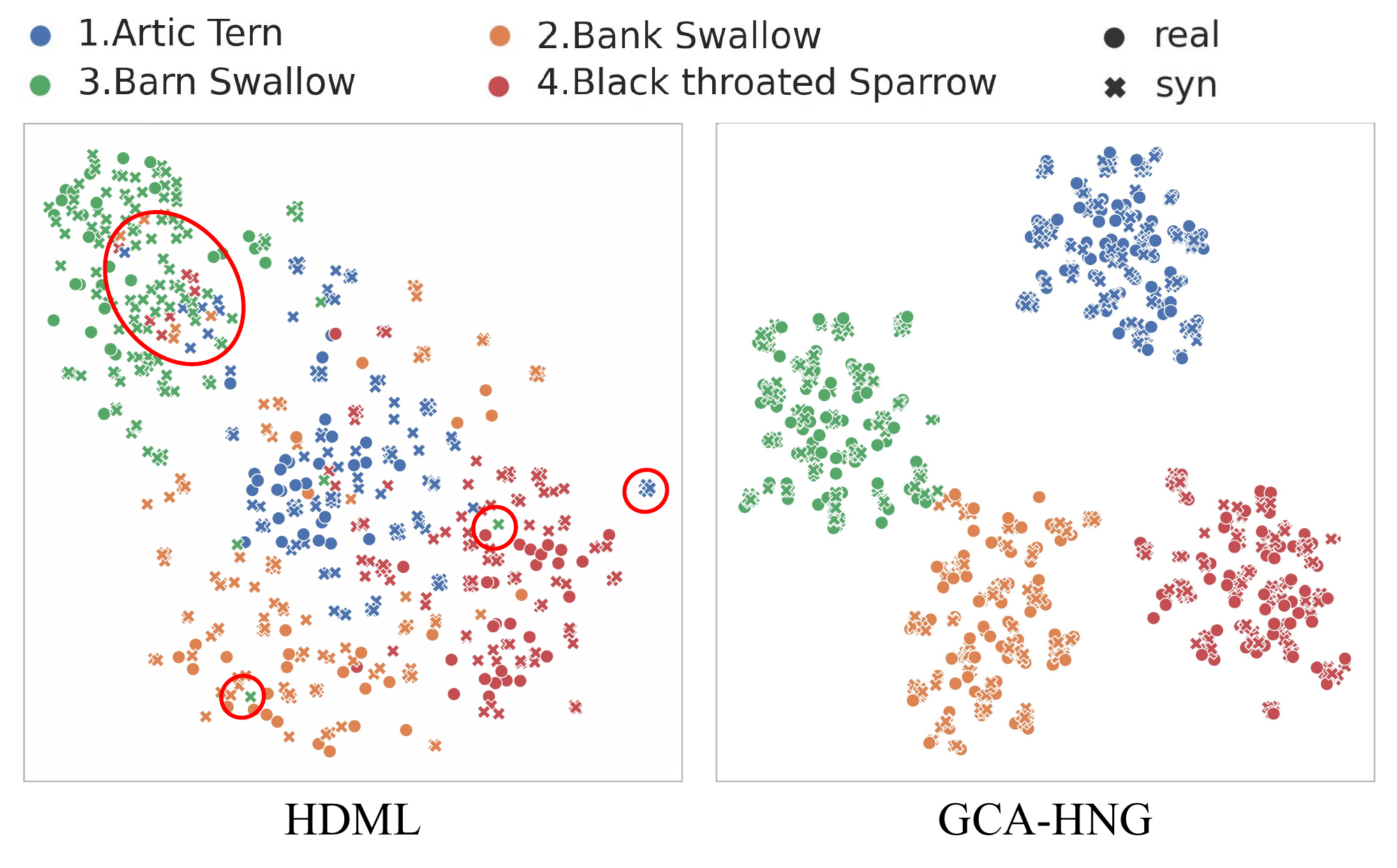}
    }
    \subfigure[Cars196]{
        \includegraphics[width=0.48\linewidth]{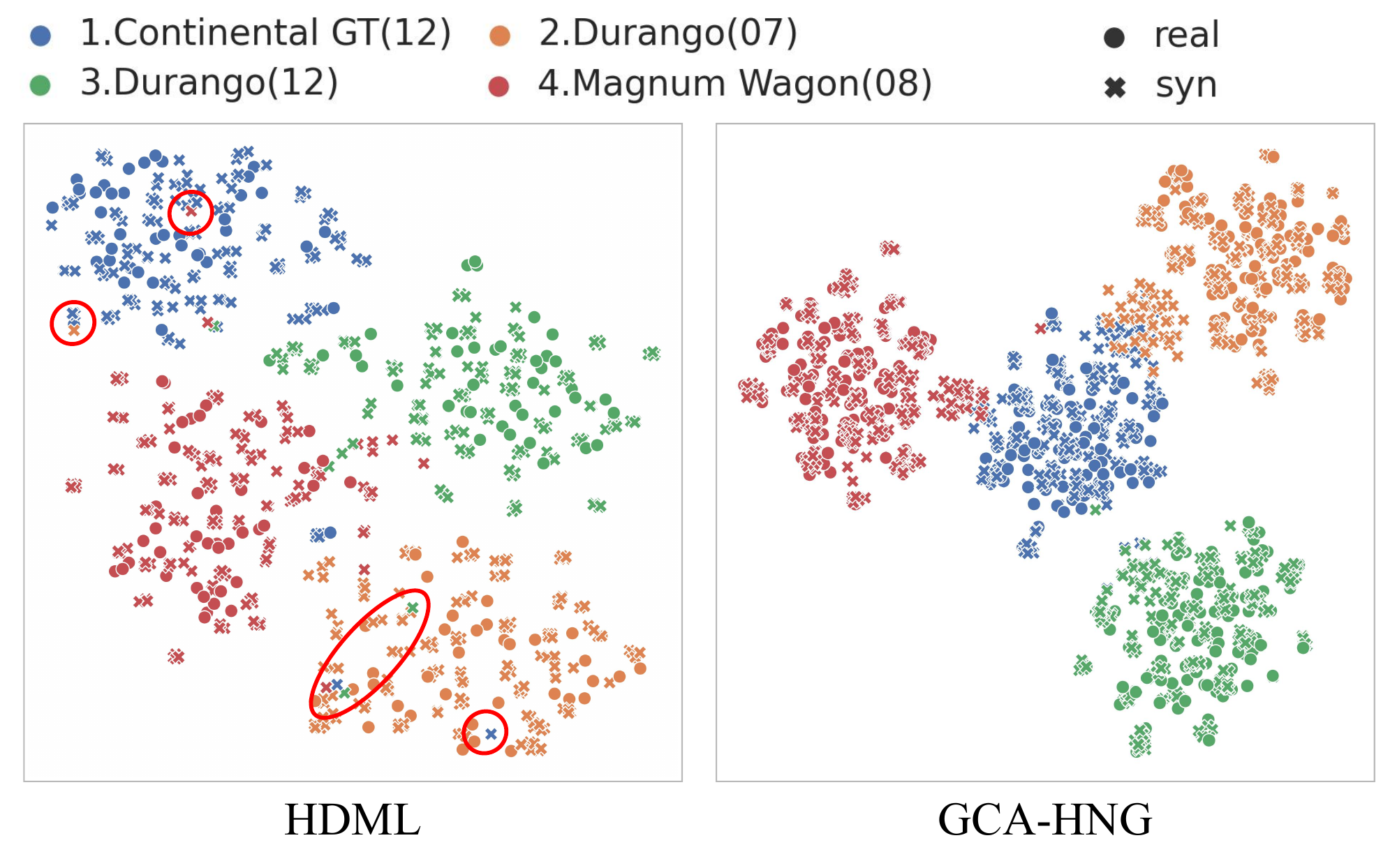}
    }
    \caption{Comparison of embedding representation distributions using t-SNE visualization for the CUB-200-2011 and Cars196 datasets after the same training iterations for HDML and our proposed GCA-HNG. The red circles present numerous noisy samples generated by HDML that disturb embedding space optimization. Our proposed GCA-HNG adaptively adjusts the hardness of the synthetic negatives to enhance their informativeness, significantly improving the convergence effect in the metric learning process.}
    \label{Fig_tsne}
\end{figure*}

To thoroughly analyze GCA-HNG performance, we further qualitatively compare it with a current typical HNG methodology and a baseline approach, examining the performance from two distinct perspectives. 

Firstly, current leading HNG methods \cite{duan2019deep,gu2020symmetrical,zhao2018adversarial,zheng2019hardness,zhu2022construct} such as HDML \cite{zheng2019hardness} typically overlook global sample correlations. This oversight results in the generation of less informative synthetic negatives, leading to inefficient network convergence. Instead, our GCA-HNG framework learns sample correlations globally and generates more informative negatives by channel-adaptive interpolation based on these correlations. For a clear comparison, we use t-SNE visualizations to showcase the differences in embedding representation distributions between HDML and our proposed GCA-HNG framework, employing ResNet-50 as the backbone and our modified NP loss for metric loss. The datasets in focus are CUB-200-2011 and Cars196. As illustrated in Fig. \ref{Fig_tsne}, under the same training iterations, the embedding representations derived from HDML reveal that the metric model shows poor class distinction. Notably, certain class pairs (class 3 and class 2/4 on CUB-200-2011, and class 1 and class 2 on Cars196) appear distant in the distribution. Despite this, HDML still generates hard negatives that are considered outliers between these classes, marked by red circles. Conversely, GCA-HNG leverages global similarity correlations with other negative classes to generate hard negatives with appropriate hardness. This approach facilitates a more effective convergence effect and fosters more discriminative metric learning, as evidenced by the t-SNE plots.

\begin{figure*}
    \centering
    \includegraphics[width=0.99\linewidth]{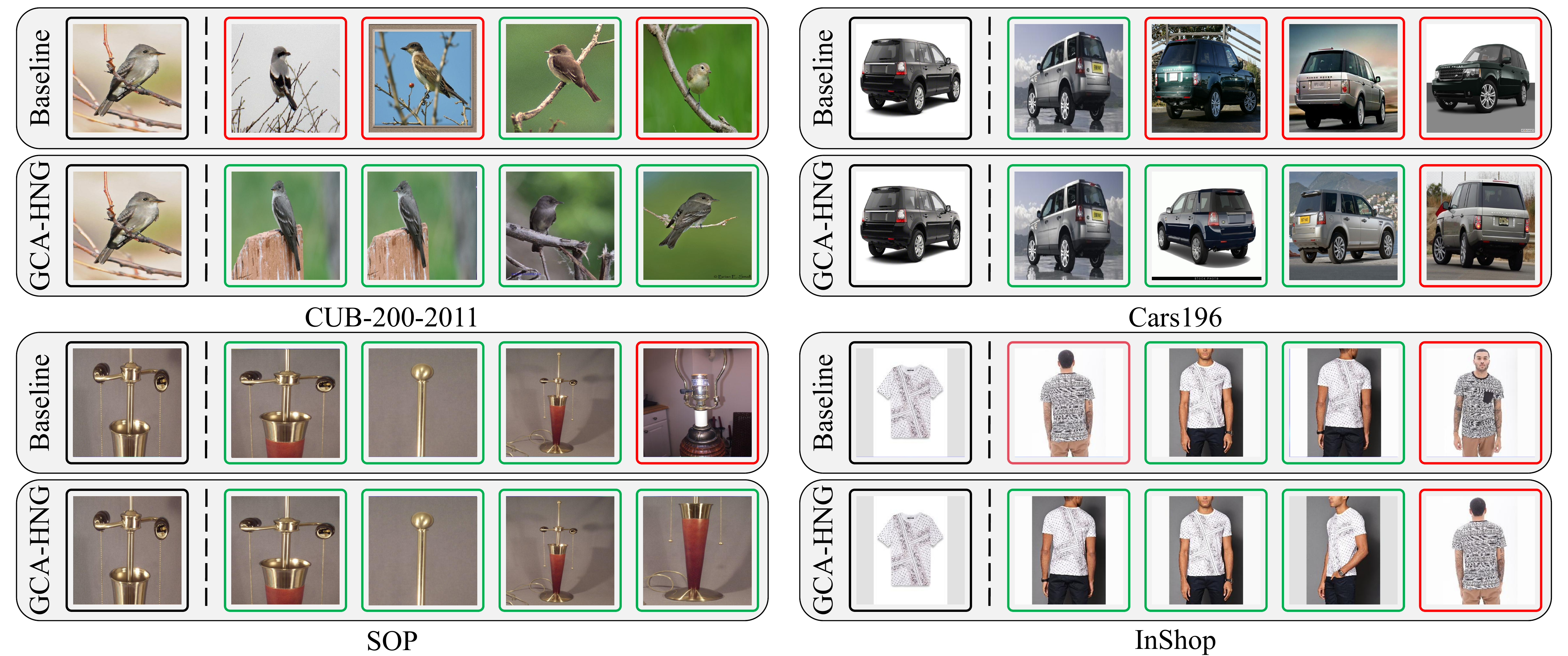}
    \caption{Comparison of retrieval results between the Baseline and GCA-HNG for four images on the CUB-200-2011, Cars196, SOP, and InShop datasets. The image to the left of the dashed line is the query, while the images to the right are Top-4 retrieval results, respectively. The green box indicates that the retrieved result belongs to the same category as the query, and the red box vice versa.}
    \label{Fig_top4}
\end{figure*}

Secondly, to intuitively illustrate how GCA-HNG enhances metric model performance, we present qualitative comparison retrieval results between the Baseline and GCA-HNG using four query images commonly misidentified by the Baseline from the CUB-200-2011, Cars196, SOP, and InShop datasets. This experiment, conducted using ResNet-50 combined with PA loss and illustrated in Fig. \ref{Fig_top4}, shows that on the CUB-200-2011 dataset, the Baseline’s first retrieved bird has a torso color similar to the query image, and its second retrieved bird has a head highly akin to the query, representing a typical hard negative. GCA-HNG, however, does not include these images in the Top-4 results, indicating its improved ability to discern subtle differences in local features. On the Cars196 dataset, the Baseline retrieves cars that closely match the query in appearance and geometry as the second and third results, yet our model effectively distinguishes these challenging samples despite their similar global visual features. Observations from the SOP and InShop datasets further confirm its robustness in handling variations in attributes such as color, scale, and viewpoint, significantly enhancing overall performance.

\subsection{Ablation Studies}
In this section, we conducted extensive ablation studies on the widely used CUB-200-2011, Cars196, and InShop datasets with the setting of ResNet-50 combined with PA and our modified NP losses to verify the robustness of GCA-HNG in various aspects comprehensively.

\subsubsection{Impact of Global Sample Correlations}
\begin{table*}
\caption{Comparison of various interpolation manners for HNG with the setting of ResNet-50 combined with PA and our modified NP losses on the CUB-200-2011, Cars196, and InShop datasets.}
\centering
\tabcolsep=1mm
\begin{tabular}{lcccc|ccc|ccc}
\toprule
 \multirow{2}{*}{Method} & \multirow{2}{*}{Loss} & \multicolumn{3}{c}{CUB-200-2011} & \multicolumn{3}{c}{Cars196} & \multicolumn{3}{c}{InShop} \\ \cmidrule{3-11}
  &  &  R@1 & RP & M@R & R@1 & RP & M@R &  R@1 & RP & M@R \\ \midrule
\midrule
single coefficient interp. & NP${}^{+}$ & 68.0$\pm$0.2 & 37.4$\pm$0.2 & 26.4$\pm$0.2 & 86.6$\pm$0.2 & 40.9$\pm$0.3 & 30.9$\pm$0.3 & 91.1$\pm$0.2 & 70.1$\pm$0.2 & 67.4$\pm$0.2 \\
w/o global correlations    & NP${}^{+}$ & 69.1$\pm$0.3 & 38.0$\pm$0.4 & 27.0$\pm$0.4 & 87.1$\pm$0.2 & 41.2$\pm$0.2 & 31.3$\pm$0.2 & 91.3$\pm$0.1 & 70.4$\pm$0.2 & 67.7$\pm$0.2 \\
w/o Hadamard product add.  & NP${}^{+}$ & 69.7$\pm$0.1 & 39.0$\pm$0.3 & 28.0$\pm$0.3 & 87.4$\pm$0.1 & 41.5$\pm$0.2 & 31.6$\pm$0.2 & 91.4$\pm$0.1 & 70.4$\pm$0.1 & 67.7$\pm$0.1 \\ 
w/o random weighting       & NP${}^{+}$ & 69.9$\pm$0.2 & 39.2$\pm$0.1 & 28.2$\pm$0.1 & 87.4$\pm$0.3 & 41.9$\pm$0.4 & 31.9$\pm$0.4 & 91.4$\pm$0.1 & 70.7$\pm$0.1 & 68.1$\pm$0.1 \\ 
GCA-HNG                    & NP${}^{+}$ & 70.3$\pm$0.3 & 39.4$\pm$0.2 & 28.4$\pm$0.3 & 88.2$\pm$0.3 &  42.3$\pm$0.4 & 32.5$\pm$0.4 & 91.7$\pm$0.1 & 71.2$\pm$0.4 & 68.5$\pm$0.3 \\ \midrule
single coefficient interp. & PA & 68.8$\pm$0.3 & 38.5$\pm$0.3 & 27.7$\pm$0.3 & 90.2$\pm$0.1 & 43.3$\pm$0.3 & 34.5$\pm$0.4 & 93.0$\pm$0.1 & 71.3$\pm$0.1 & 68.8$\pm$0.1 \\ 
w/o global correlations    & PA & 69.3$\pm$0.3 & 39.0$\pm$0.2 & 28.1$\pm$0.2 & 90.5$\pm$0.1 & 43.6$\pm$0.3 & 34.8$\pm$0.4 & 93.2$\pm$0.1 & 71.4$\pm$0.2 & 69.0$\pm$0.2 \\
w/o Hadamard product add.  & PA & 71.7$\pm$0.2 & 40.3$\pm$0.3 & 29.7$\pm$0.3 & 90.7$\pm$0.1 & 44.0$\pm$0.1 & 35.1$\pm$0.1 & 93.2$\pm$0.2 & 71.5$\pm$0.2 & 69.1$\pm$0.1 \\
w/o random weighting       & PA & 71.3$\pm$0.2 & 40.2$\pm$0.3 & 29.4$\pm$0.4 & 90.7$\pm$0.1 & 43.8$\pm$0.1 & 35.0$\pm$0.1 & 93.3$\pm$0.1 & 71.5$\pm$0.2 & 69.0$\pm$0.1 \\
GCA-HNG                    & PA & 72.3$\pm$0.3 & 40.7$\pm$0.3 & 30.1$\pm$0.3 & 91.1$\pm$0.2 & 44.1$\pm$0.3 & 35.4$\pm$0.3 & 93.6$\pm$0.1 & 71.9$\pm$0.2 & 69.5$\pm$0.2 \\ \bottomrule
\end{tabular}
\label{table_interpolation manner}
\end{table*}

To quantitatively assess the effectiveness of global sample correlations in the GCA-HNG framework, we conducted a comparative analysis between the complete GCA-HNG implementation and a variant lacking these correlations, termed as ``w/o global correlations''. This version omits the node message propagation in GCA-HNG, retaining solely the edge message propagation, which restricts each edge to information from just its two adjacent samples, thereby precluding access to the broader global information within the entire graph structure. The experiment results, presented in Table \ref{table_interpolation manner}, demonstrate that under both PA and our modified NP loss settings on the CUB-200-2011, Cars196, and InShop datasets, the GCA-HNG with global sample correlations yields a significant enhancement across all retrieval metrics. These results substantiate the pivotal role of node message propagation in modeling global sample correlations and effectively guiding the generation of hard negatives.

Further analysis delved into the influence of edge-to-node information fusion within the node message propagation. Specifically, we examined a scenario wherein omits the Hadamard product addition $E_{ij}$ from the model, designated as ``w/o Hadamard product add.''. In this setup, each node interacts only with its neighboring nodes, neglecting the edge information. Table \ref{table_interpolation manner} indicates that GCA-HNG presents moderate performance enhancements for both loss functions across three datasets. This observation confirms the efficacy of incorporating the Hadamard product addition $E_{ij}$ in the node message propagation mechanism for learning global sample correlations.

\subsubsection{Analysis of CACAI Module}
To evaluate the effectiveness of our proposed CACAI approach in utilizing correlation information for generating hard negatives, we conducted ablation experiments comparing three interpolation methods, detailed in Table \ref{table_interpolation manner}. The ``single coefficient interp.'' method utilizes a uniform interpolation coefficient across all channels, derived from local correlations between two samples, similar to the approach in HDML \cite{zheng2019hardness} with $\bm{\lambda}$ set to 1. While the ``w/o global correlations" variant emphasizes learning channel-specific local correlation representations between sample pairs, assigning adaptive coefficients to each channel without incorporating global correlations. The third method, ``w/o random weighting'', involves the arbitrary selection of an interpolated representation from the set $\{\tilde{\textbf{z}}_{i j}|l_j=n\}$ rather than combining them to form a synthetic negative, maintaining a consistent count of synthetic negatives for $J_{syn}$. 
Experiment results across various loss functions and datasets demonstrate that ``w/o global correlations" exceeds ``single coefficient interp." in performance, and GCA-HNG outperforms the ``w/o random weighting'' method. These findings affirm the effectiveness of correlation-aware channel-level embedding interpolation and random weighting fusion, respectively. 

\subsubsection{Impact of Class Perceptual Scales on Global Sample Correlations}
\label{class_sampling_sec}
\begin{figure}
    \centering
    \subfigure[CUB-200 with NP${}^{+}$ loss]{
        \includegraphics[width=0.45\linewidth]{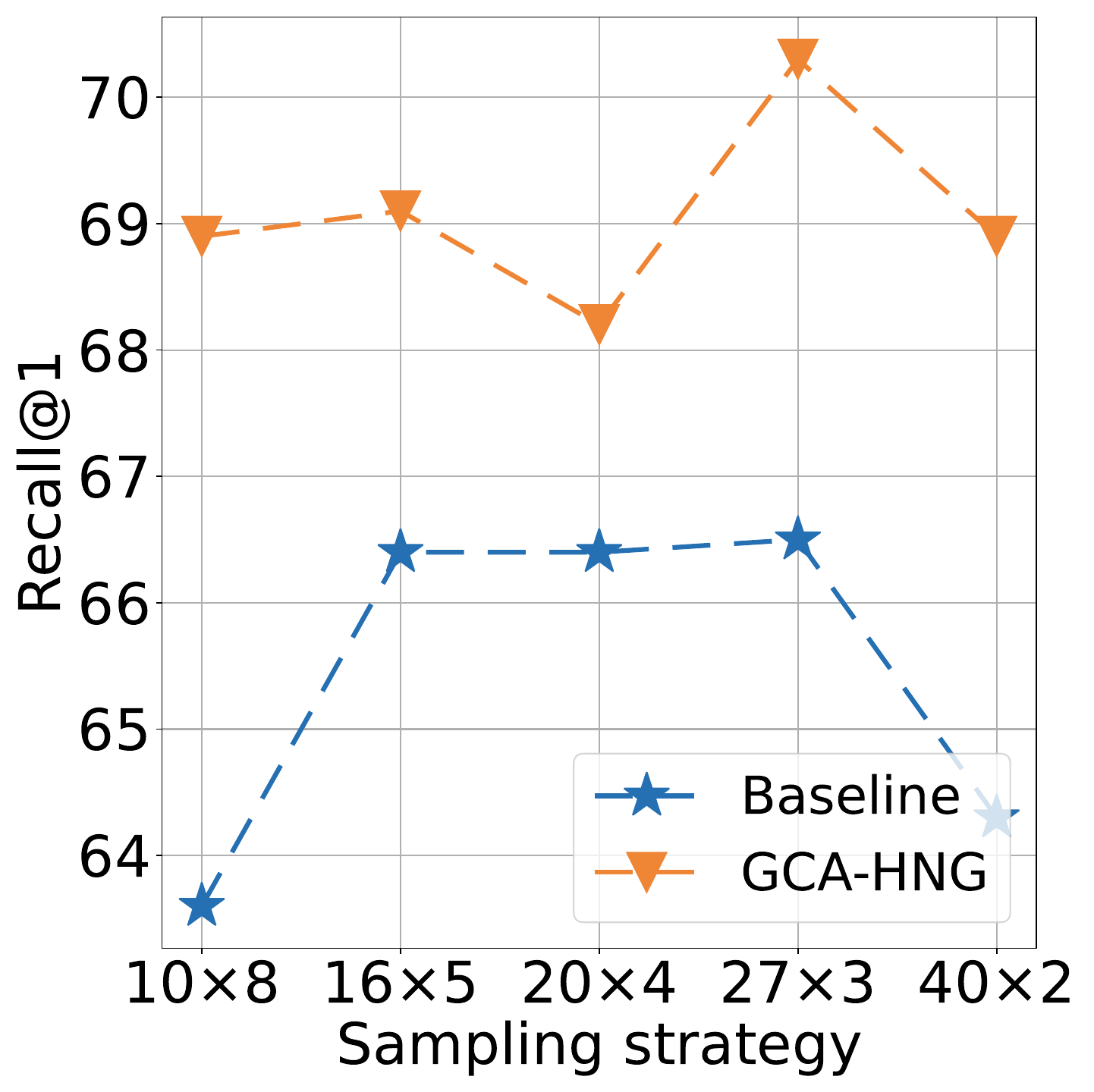}
    }
    \subfigure[Cars196 with NP${}^{+}$ loss]{
        \includegraphics[width=0.45\linewidth]{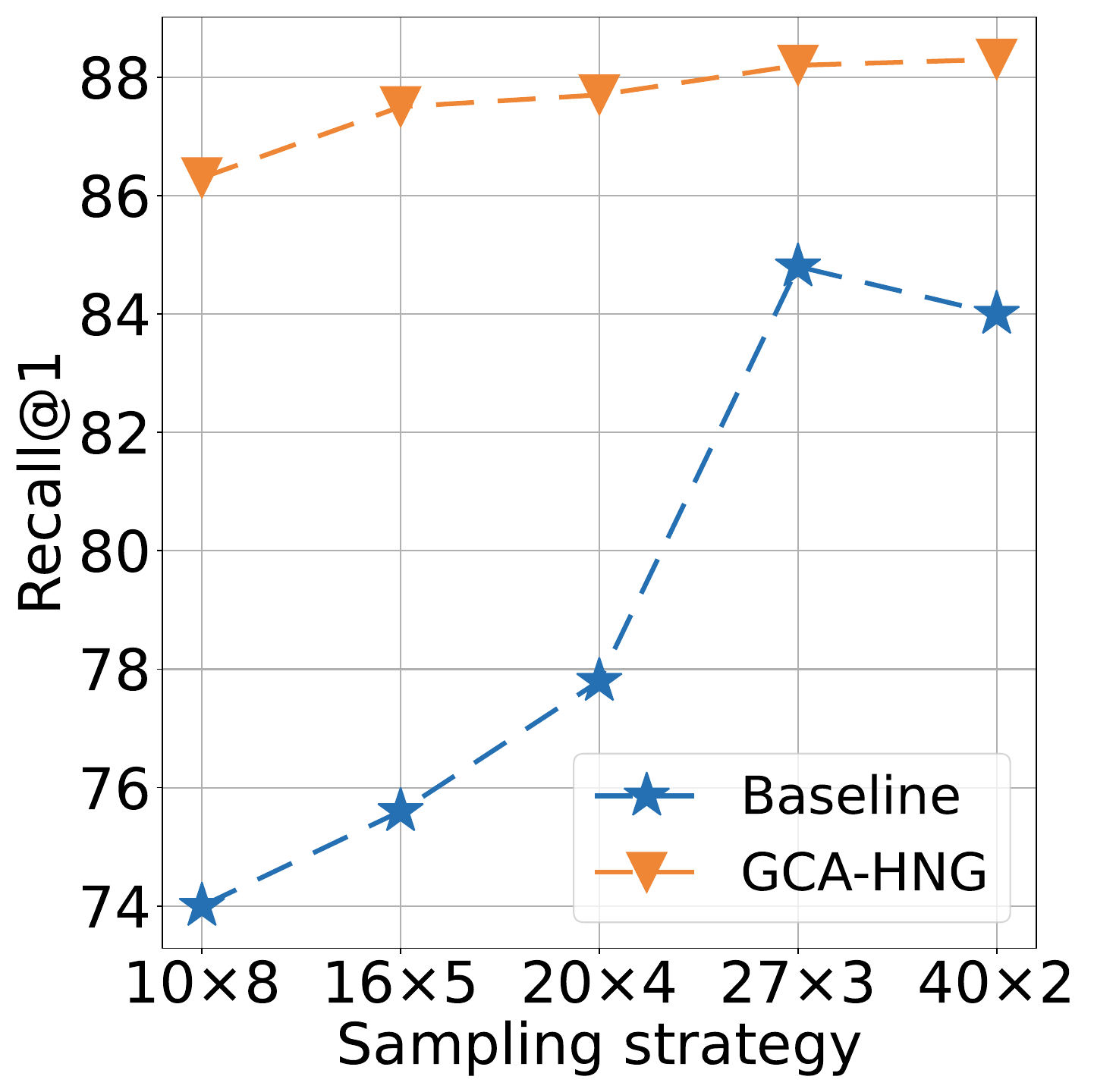}
    }
    \\
    \subfigure[CUB-200 with PA loss]{
        \includegraphics[width=0.45\linewidth]{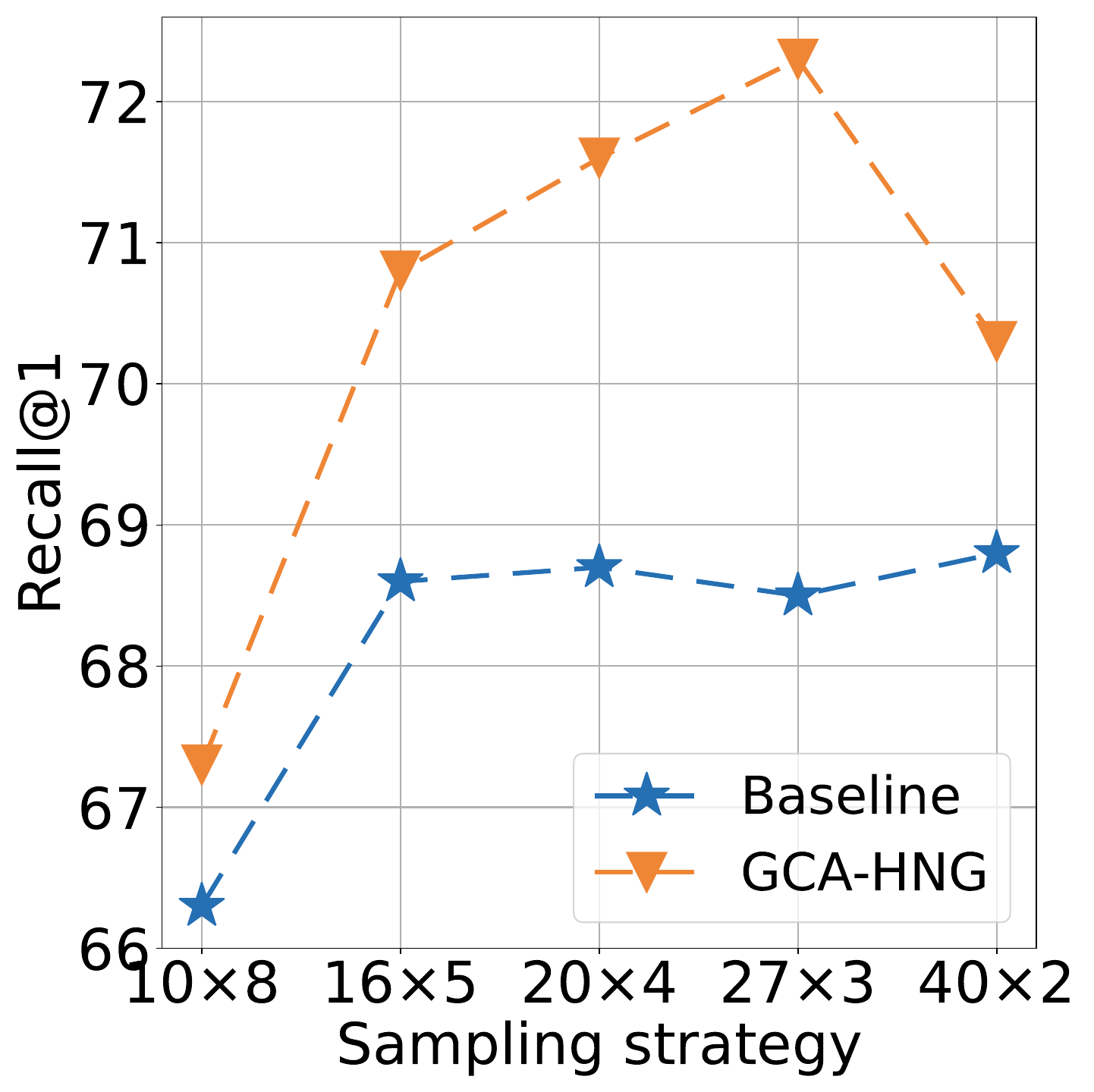}
    }
    \subfigure[Cars196 with PA loss]{
        \includegraphics[width=0.45\linewidth]{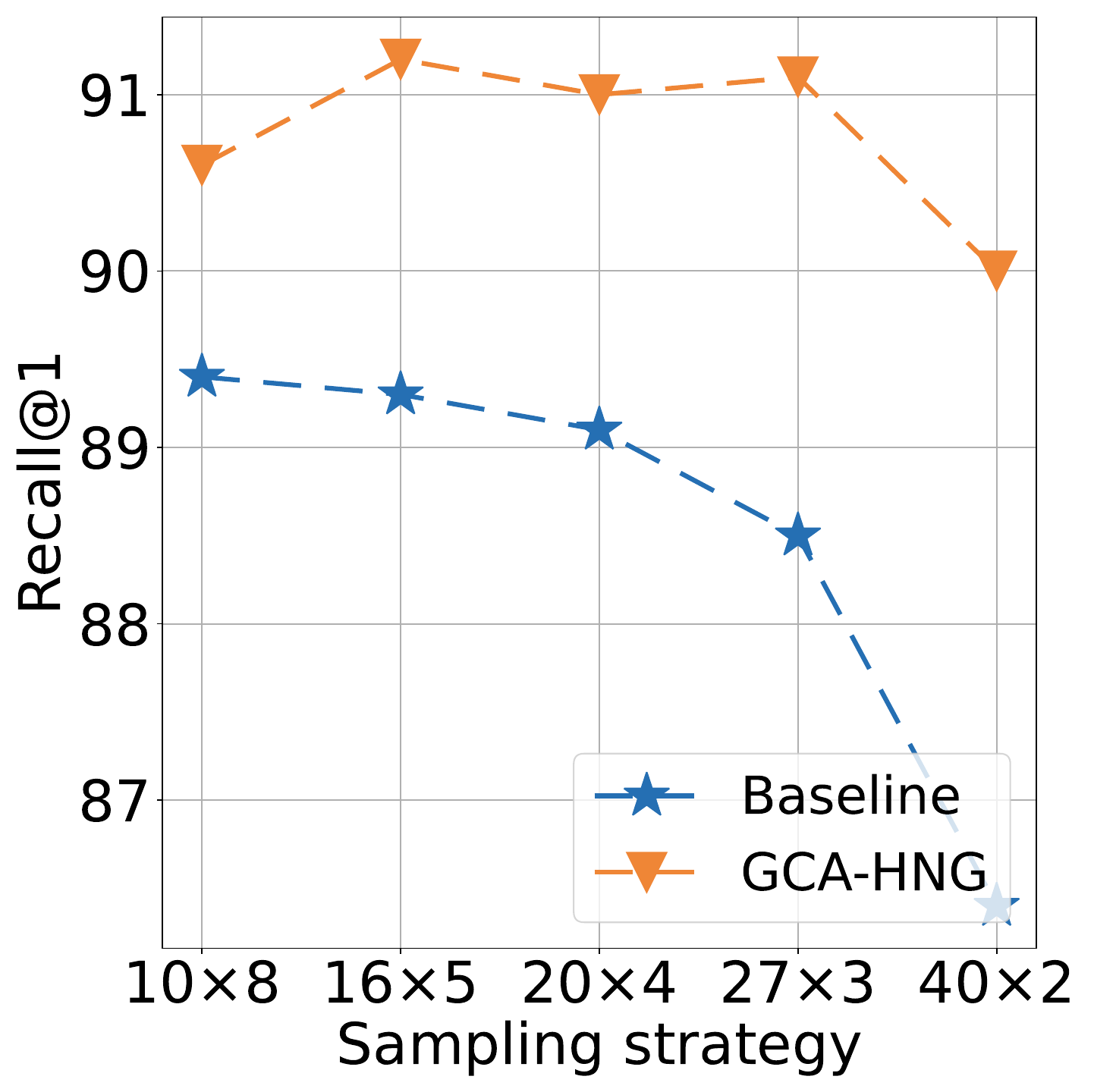}
    }
    \caption{Comparison of Recall@1 results versus different class perceptual scales with the setting of ResNet-50 combined with PA and our modified NP losses on the CUB-200-2011 and Cars196 datasets.}
    \label{Fig_batch size}
\end{figure}

To investigate the impact of class perceptual scales on global sample correlations, we conducted targeted experiments focusing on how varying the number of sampled classes within the mini-batch affects the performance of GCA-HNG. Employing a balanced sampling strategy denoted as $\mathcal{N} \times m$, we sampled $\mathcal{N}$ distinct classes, each with $m$ instances, maintaining a batch size of about 80. We tested configurations including $10 \times 8$, $16 \times 5$, $20 \times 4$, $27 \times 3$, and $40 \times 2$. To mitigate potential biases introduced by the balanced sampling strategy on the PA and our modified NP losses, we incorporated baseline settings with an identical sampling strategy for a fair comparison. This study focused on the CUB-200-2011 and Cars196 datasets, excluding InShop due to its many classes with fewer than three samples. As shown in Fig. \ref{Fig_batch size}, GCA-HNG exceeds the baseline remarkably across all configurations, demonstrating strong robustness against variation in sampling strategies. Particularly with the Cars196 dataset using a $10 \times 8$ strategy under the our modified NP loss setting, GCA-HNG demonstrates an impressive enhancement, surpassing the baseline by over $12\%$ on Recall@1. The trend delineated by the GCA-HNG's performance, depicted as a folded line, indicates a trade-off between the quantity of sampled classes and model performance. Based on these results, we adopt a 27$\times$3 sampling strategy for both datasets, as it offers competitive performance.

\subsubsection{Impact of Losses for GCA-HNG}
During the first stage of optimization, the loss $J_{gen}$, as described in Eq.(\ref{eq_Jgen}), plays a crucial role in guiding the generation of informative negatives. To evaluate the individual contributions of each component within this loss function, we take turns omitting one loss component at a time and analyze its impact on model performance. Table \ref{table_loss} indicates that each loss component distinctly influences the outcomes. In the PA loss setting, $J_{ce}$ exhibits a remarkable effect, while in our modified NP loss setting, all three losses: $J_{ce}$, $J_{sim}$, and $J_{div}$, prove essential. $J_{ce}$ ensures synthetic negatives preserve their class representations, aiding the model in learning accurate class boundaries. $J_{sim}$ and $J_{div}$ improve model performance by reasonably enhancing the hardness and diversity of synthetic negatives, respectively. An integration of these three losses yields a better performance.

Furthermore, we conducted an exhaustive analysis by varying the weights of each loss component within $J_{gen}$. Table \ref{table_hyper-parameters} presents the ablation results of hyper-parameters $\gamma_{s}$ and $\gamma_{d}$, under the configuration of ResNet-50 combined with PA and our modified NP losses on the CUB-200-2011, Cars196, and InShop datasets. In this experiment, we varied the values of $\gamma_s$ or $\gamma_d$ by increasing them tenfold or reducing them to a tenth to verify their impact on performance. Results from three datasets confirm that the existing hyper-parameters are optimal, exhibiting superior model robustness over a wide range of parameter variations, especially in the PA loss setting.

\begin{table*}
\caption{Ablation results of each loss component for the loss $J_{gen}$ with the setting of ResNet-50 combined with PA and our modified NP losses on the CUB-200-2011, Cars196, and InShop datasets.}
\centering
\tabcolsep=1.5mm
\begin{tabular}{lcccccc|ccc|ccc}
\toprule
 \multirow{2}{*}{$J_{ce}$} & \multirow{2}{*}{$J_{sim}$} & \multirow{2}{*}{$J_{div}$} & \multirow{2}{*}{Loss} & \multicolumn{3}{c}{CUB-200-2011} & \multicolumn{3}{c}{Cars196} & \multicolumn{3}{c}{InShop}\\ \cmidrule{5-13}
& & & & R@1 & RP & M@R & R@1 & RP & M@R & R@1 & RP & M@R \\ \midrule
\midrule
& \checkmark & \checkmark &             NP${}^{+}$ & 67.7$\pm$0.3 & 36.8$\pm$0.4 & 25.9$\pm$0.4 & 86.7$\pm$0.2 & 41.4$\pm$0.1 & 31.4$\pm$0.2 & 91.3$\pm$0.1 & 70.7$\pm$0.1 & 68.1$\pm$0.2 \\
\checkmark & & \checkmark &             NP${}^{+}$ & 67.8$\pm$0.2 & 37.4$\pm$0.1 & 26.4$\pm$0.2 & 87.0$\pm$0.2 & 41.3$\pm$0.2 & 31.3$\pm$0.2 & 91.4$\pm$0.1 & 70.3$\pm$0.2 & 67.6$\pm$0.1 \\
\checkmark & \checkmark & &             NP${}^{+}$ & 67.8$\pm$0.3 & 37.4$\pm$0.3 & 26.4$\pm$0.3 & 86.8$\pm$0.2 & 41.1$\pm$0.1 & 31.2$\pm$0.2 & 91.1$\pm$0.1 & 70.2$\pm$0.1 & 67.5$\pm$0.1 \\
\checkmark & \checkmark & \checkmark &  NP${}^{+}$ & 70.3$\pm$0.3 & 39.4$\pm$0.2 & 28.4$\pm$0.3 & 88.2$\pm$0.3 & 42.3$\pm$0.4 & 32.5$\pm$0.4 & 91.7$\pm$0.1 & 71.2$\pm$0.4 & 68.5$\pm$0.3 \\ \midrule
& \checkmark & \checkmark &             PA & 70.1$\pm$0.1 & 39.3$\pm$0.1 & 28.6$\pm$0.1 & 90.5$\pm$0.1 & 43.4$\pm$0.1 & 34.5$\pm$0.1 & 93.1$\pm$0.1 & 71.7$\pm$0.1 & 69.3$\pm$0.1 \\
\checkmark & & \checkmark &             PA & 71.3$\pm$0.1 & 40.1$\pm$0.2 & 29.3$\pm$0.2 & 90.6$\pm$0.1 & 43.7$\pm$0.1 & 34.7$\pm$0.1 & 93.0$\pm$0.1 & 71.5$\pm$0.2 & 69.1$\pm$0.1 \\
\checkmark & \checkmark & &             PA & 71.4$\pm$0.3 & 40.0$\pm$0.2 & 29.3$\pm$0.2 & 90.8$\pm$0.1 & 43.8$\pm$0.1 & 35.0$\pm$0.1 & 93.2$\pm$0.1 & 71.6$\pm$0.1 & 69.2$\pm$0.1 \\
\checkmark & \checkmark & \checkmark &  PA & 72.3$\pm$0.3 & 40.7$\pm$0.3 & 30.1$\pm$0.3 & 91.1$\pm$0.2 & 44.1$\pm$0.3 & 35.4$\pm$0.3 & 93.6$\pm$0.1 & 71.9$\pm$0.2 & 69.5$\pm$0.2 \\ \bottomrule
\end{tabular}
\label{table_loss}
\end{table*}

\begin{table*}
\caption{Ablation results of the hyper-parameters $\gamma_{s}$ and $\gamma_{d}$ for the loss $J_{gen}$ with the setting of ResNet-50 combined with PA and our modified NP losses on the CUB-200-2011, Cars196, and InShop datasets.}
\centering
\tabcolsep=2mm
\begin{tabular}{lccccc|ccc|ccc}
\toprule
 \multirow{2}{*}{$\gamma_{s}$} & \multirow{2}{*}{$\gamma_{d}$} & \multirow{2}{*}{Loss} & \multicolumn{3}{c}{CUB-200-2011} & \multicolumn{3}{c}{Cars196} & \multicolumn{3}{c}{InShop} \\ \cmidrule{4-12}
& & & R@1 & RP & M@R & R@1 & RP & M@R & R@1 & RP & M@R \\ \midrule \midrule
0.1  & 0.03  & NP${}^{+}$ & 68.4$\pm$0.2 & 37.7$\pm$0.3 & 26.8$\pm$0.4 & 87.2$\pm$0.3 & 41.6$\pm$0.2 & 31.5$\pm$0.1 & 91.3$\pm$0.1 & 70.3$\pm$0.2 & 67.7$\pm$0.2 \\
10.0 & 0.03  & NP${}^{+}$ & 67.8$\pm$0.2 & 37.1$\pm$0.2 & 26.1$\pm$0.2 & 87.3$\pm$0.3 & 41.3$\pm$0.4 & 31.4$\pm$0.3 & 91.4$\pm$0.1 & 70.4$\pm$0.1 & 67.7$\pm$0.1 \\
1.0  & 0.003 & NP${}^{+}$ & 68.7$\pm$0.3 & 37.9$\pm$0.4 & 27.0$\pm$0.4 & 87.1$\pm$0.2 & 41.3$\pm$0.2 & 31.4$\pm$0.3 & 91.2$\pm$0.1 & 70.2$\pm$0.1 & 67.5$\pm$0.1 \\
1.0  & 0.3   & NP${}^{+}$ & 68.6$\pm$0.3 & 37.9$\pm$0.4 & 26.9$\pm$0.4 & 87.2$\pm$0.2 & 41.7$\pm$0.4 & 31.5$\pm$0.2 & 91.3$\pm$0.1 & 70.4$\pm$0.2 & 67.7$\pm$0.1 \\ 
1.0  & 0.03  & NP${}^{+}$ & 70.3$\pm$0.3 & 39.4$\pm$0.2 & 28.4$\pm$0.3 & 88.2$\pm$0.3 & 42.3$\pm$0.4 & 32.5$\pm$0.4 & 91.7$\pm$0.1 & 71.2$\pm$0.4 & 68.5$\pm$0.3  \\ \midrule
0.1  & 0.01  & PA & 71.6$\pm$0.1 & 40.2$\pm$0.1 & 29.5$\pm$0.1 & 90.7$\pm$0.2 & 44.1$\pm$0.3 & 35.3$\pm$0.3 & 93.1$\pm$0.1 & 71.6$\pm$0.2 & 69.2$\pm$0.2 \\ 
10.0 & 0.01  & PA & 71.6$\pm$0.3 & 40.4$\pm$0.3 & 29.7$\pm$0.3 & 90.4$\pm$0.1 & 43.4$\pm$0.3 & 34.4$\pm$0.3 & 93.0$\pm$0.1 & 71.4$\pm$0.2 & 68.9$\pm$0.1 \\ 
1.0  & 0.001 & PA & 71.8$\pm$0.2 & 39.9$\pm$0.2 & 29.1$\pm$0.2 & 90.9$\pm$0.1 & 43.5$\pm$0.1 & 34.7$\pm$0.1 & 93.3$\pm$0.1 & 71.6$\pm$0.2 & 69.2$\pm$0.2 \\ 
1.0  & 0.1   & PA & 70.9$\pm$0.2 & 40.0$\pm$0.4 & 29.2$\pm$0.4 & 90.7$\pm$0.2 & 43.9$\pm$0.2 & 35.2$\pm$0.2 & 93.0$\pm$0.1 & 71.4$\pm$0.1 & 68.9$\pm$0.1 \\ 
1.0  & 0.01  & PA & 72.3$\pm$0.3 & 40.7$\pm$0.3 & 30.1$\pm$0.3 & 91.1$\pm$0.2 & 44.1$\pm$0.3 & 35.4$\pm$0.3 & 93.6$\pm$0.1 & 71.9$\pm$0.2 & 69.5$\pm$0.2 \\ \bottomrule
\end{tabular}
\label{table_hyper-parameters}
\end{table*}

\subsubsection{Impact of GNN on Metric Model}
\begin{table*}
\caption{Ablation results of the proposed GNN on the performance of the metric model with the setting of ResNet-50 combined with PA and our modified NP losses on the CUB-200-2011, Cars196, and InShop datasets.}
\centering
\tabcolsep=1.7mm
\begin{tabular}{lcccc|ccc|ccc}
\toprule
\multirow{2}{*}{Model} & \multirow{2}{*}{Loss} & \multicolumn{3}{c}{CUB-200-2011} & \multicolumn{3}{c}{Cars196} & \multicolumn{3}{c}{InShop} \\ \cmidrule{3-11}
& & R@1 & RP & M@R & R@1 & RP & M@R & R@1 & RP & M@R \\ \midrule \midrule
Baseline        & NP${}^{+}$ & 66.5$\pm$0.1 & 36.5$\pm$0.3 & 25.5$\pm$0.3 & 84.8$\pm$0.1 & 39.5$\pm$0.1 & 29.3$\pm$0.1 & 89.6$\pm$0.1 & 67.6$\pm$0.2 & 64.7$\pm$0.2 \\
Baseline + GNN  & NP${}^{+}$ & 66.8$\pm$0.1 & 36.8$\pm$0.4 & 25.9$\pm$0.4 & 84.7$\pm$0.3 & 39.9$\pm$0.3 & 29.2$\pm$0.2 & 90.3$\pm$0.1 & 68.6$\pm$0.2 & 65.8$\pm$0.2 \\
GCA-HNG         & NP${}^{+}$ & 70.3$\pm$0.3 & 39.4$\pm$0.2 & 28.4$\pm$0.3 & 88.2$\pm$0.3 & 42.3$\pm$0.4 & 32.5$\pm$0.4 & 91.7$\pm$0.1 & 71.2$\pm$0.4 & 68.5$\pm$0.3 \\ \midrule
Baseline        & PA & 68.5$\pm$0.3 & 37.9$\pm$0.4 & 27.2$\pm$0.4 & 88.4$\pm$0.1 & 40.3$\pm$0.2 & 30.9$\pm$0.2 & 92.2$\pm$0.1 & 69.8$\pm$0.1 & 67.2$\pm$0.1 \\
Baseline + GNN  & PA & 68.5$\pm$0.2 & 38.5$\pm$0.2 & 27.7$\pm$0.2 & 88.3$\pm$0.1 & 40.6$\pm$0.1 & 31.0$\pm$0.1 & 92.6$\pm$0.1 & 69.9$\pm$0.2 & 67.4$\pm$0.2 \\
GCA-HNG         & PA & 72.3$\pm$0.3 & 40.7$\pm$0.3 & 30.1$\pm$0.3 & 91.1$\pm$0.2 & 44.1$\pm$0.3 & 35.4$\pm$0.3 & 93.6$\pm$0.1 & 71.6$\pm$0.2 & 69.2$\pm$0.2 \\ \bottomrule
\end{tabular}
\label{table_gnn}
\end{table*}

\begin{table}
    \caption{Comparisons of the training and inference time per iteration between the Baseline and GCA-HNG with the setting of a mini-batch of 80 and a ResNet-50 backbone network.}
    \centering
    \tabcolsep=2mm
    \begin{tabular}{lc|cc}
    \toprule
    Mode    & Loss    & Baseline & GCA-HNG\\ \midrule \midrule
    Training    & NP${}^{+}$  & 0.22 sec/iter  & 0.29 sec/iter  \\ 
    Inference   & NP${}^{+}$  & 0.09 sec/iter  & 0.09 sec/iter  \\ \midrule
    Training    & PA  & 0.21 sec/iter  & 0.29 sec/iter  \\ 
    Inference   & PA  & 0.09 sec/iter  & 0.09 sec/iter  \\
    \bottomrule
    \end{tabular}
    \label{table_time}
\end{table}

The graph network $G$ is the core module of the GCA-HNG framework, and we conducted two sets of targeted experiments focusing on its impact on the metric model.

Firstly, we conducted ablation studies to ascertain whether integrating a GNN offers additional enhancements to the metric model. The ``Baseline'' method utilized only the metric loss $J_r(\textbf{z})$ to optimize the metric model without other modules introduced. The ``Baseline+GNN'' method applied both $J_r(\textbf{z})$ and $J_{gca}$ losses from Eq.(\ref{eq_Jm}) to optimize the metric model, incorporating the proposed GNN but omitting the HNG process. Results in Table \ref{table_gnn} reflect that merely introducing a GNN alone does not significantly alter retrieval performance on the CUB-200-2011 and Cars196 datasets, yet a slight improvement on the InShop dataset. These results illustrate the substantial improvements derived from synthetic hard negatives. 

Secondly, we executed a comparative analysis to discern the effects of the GNN on the training and inference durations per iteration under PA and our modified NP loss settings. Table \ref{table_time} shows that while introducing the GNN module to model the correlations does extend the training time per iteration, it is imperative to note that the inference time remains unaffected. The GNN module assists the metric model in learning a robust embedding space in the training phase, whereas during inference, the metric model alone is utilized to obtain sample representations for similarity measurement. In practical applications, the training process is typically a one-time process, after which the trained model is deployed for extended periods of inference. Thus, the increased training time is generally not a significant concern in most applications.

\subsubsection{GNN Structure Exploration}
\label{KH_exploration_sec}
\begin{figure*}
    \centering
    \subfigure[CUB-200 with NP${}^{+}$ loss]{
        \includegraphics[width=0.25\linewidth]{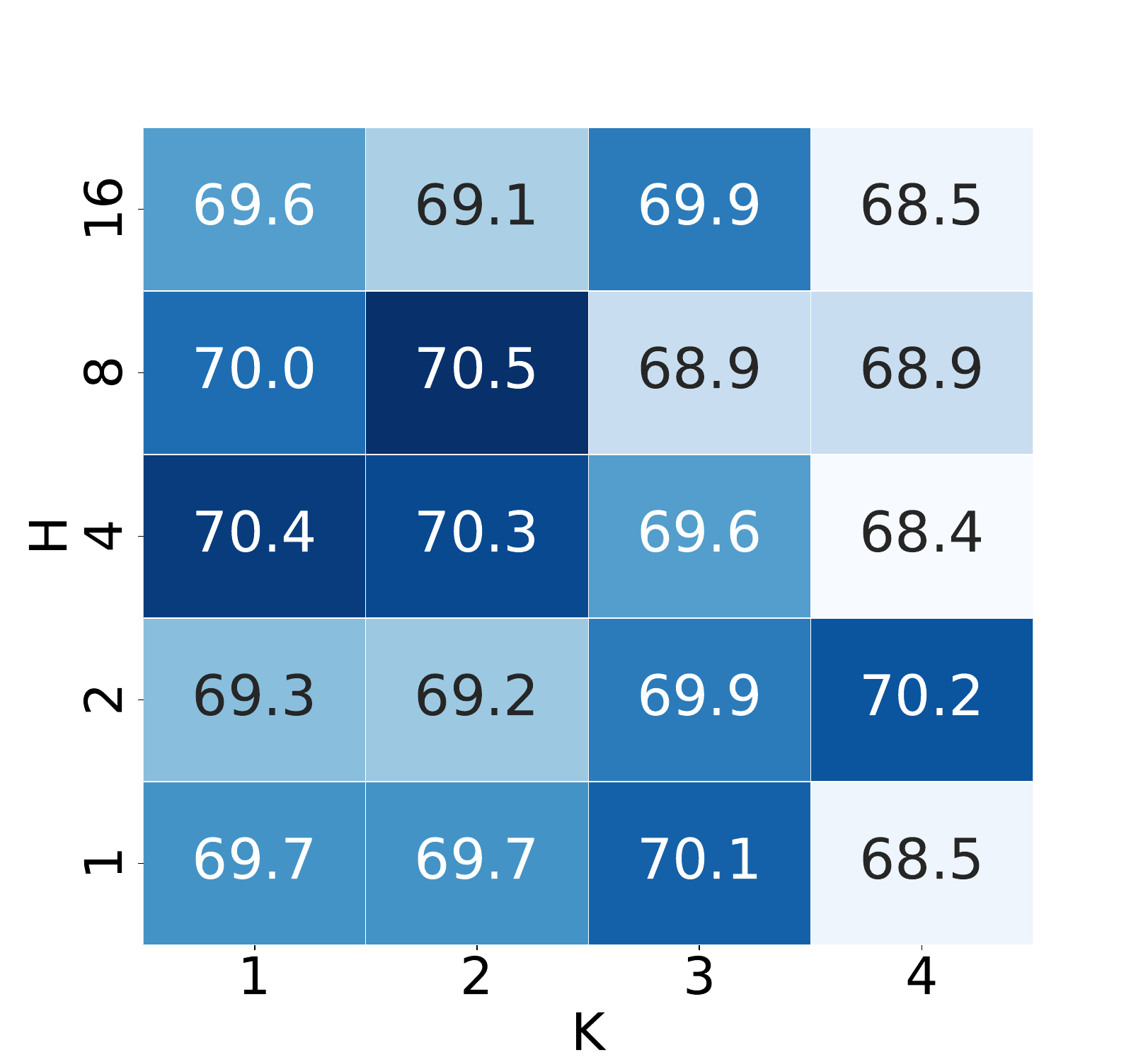}
    }
    \subfigure[Cars196 with NP${}^{+}$ loss]{
        \includegraphics[width=0.25\linewidth]{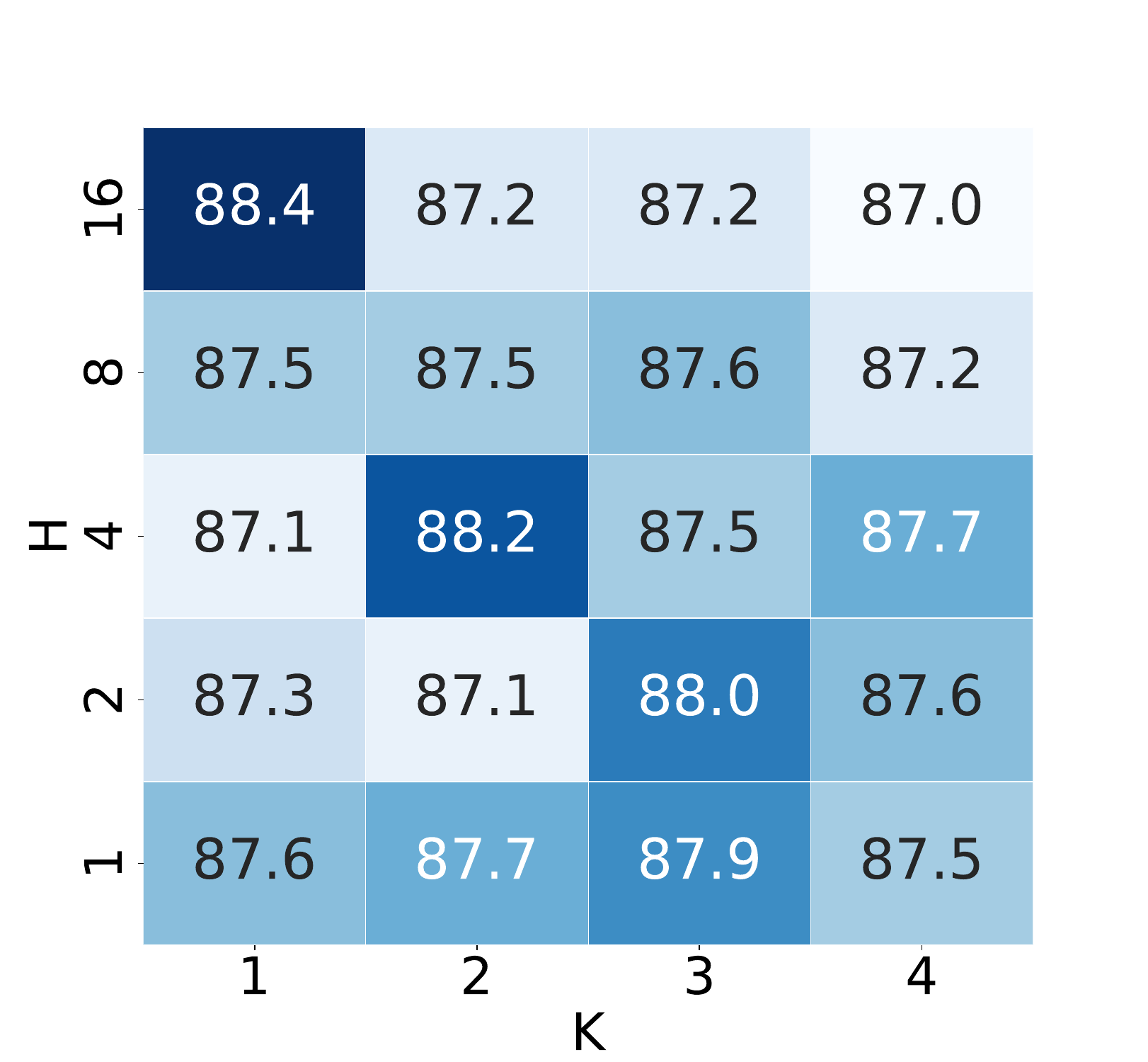}
    }
    \subfigure[InShop with NP${}^{+}$ loss]{
        \includegraphics[width=0.25\linewidth]{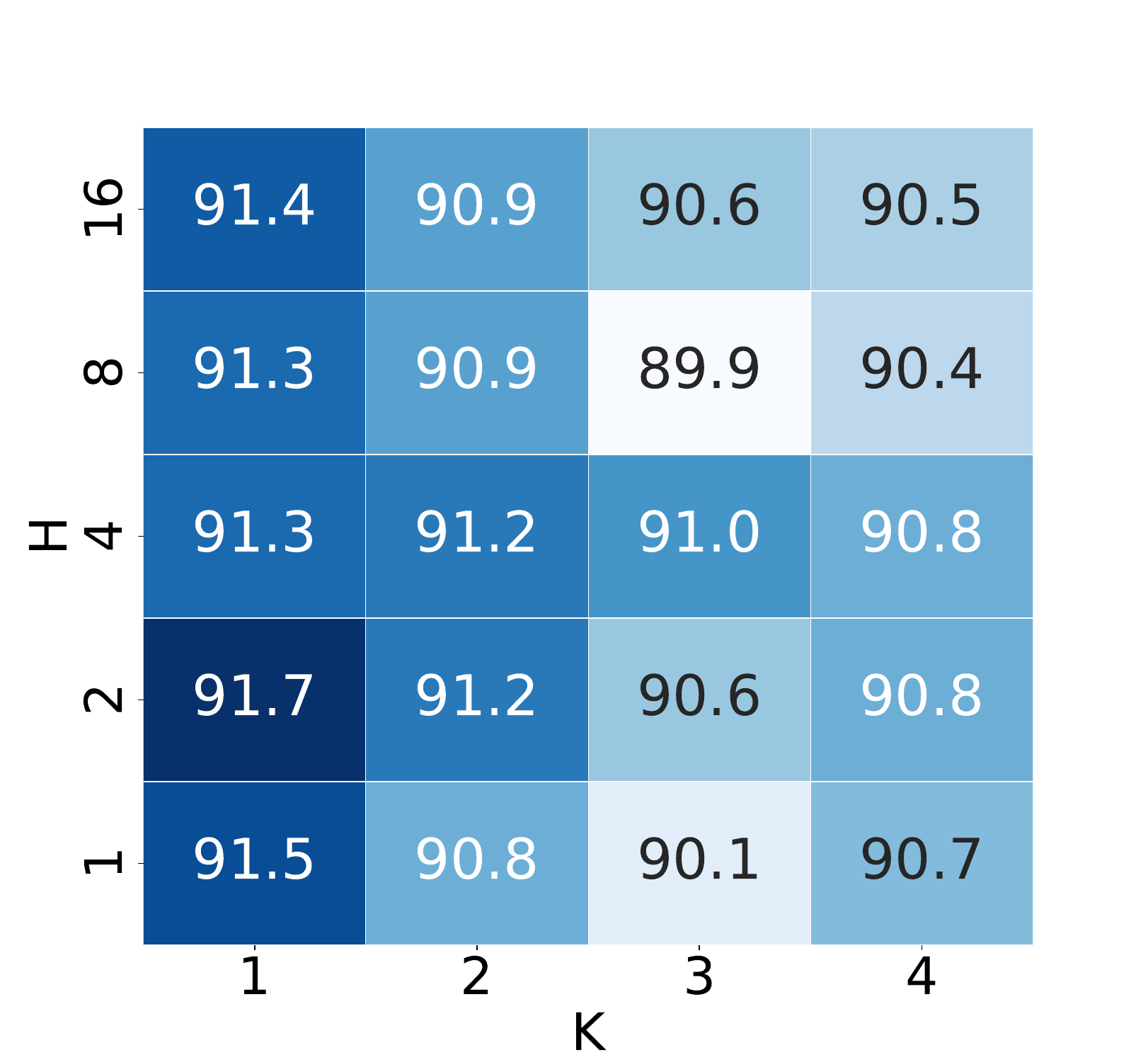}
    }
    \\
    \subfigure[CUB-200 with PA loss]{
        \includegraphics[width=0.25\linewidth]{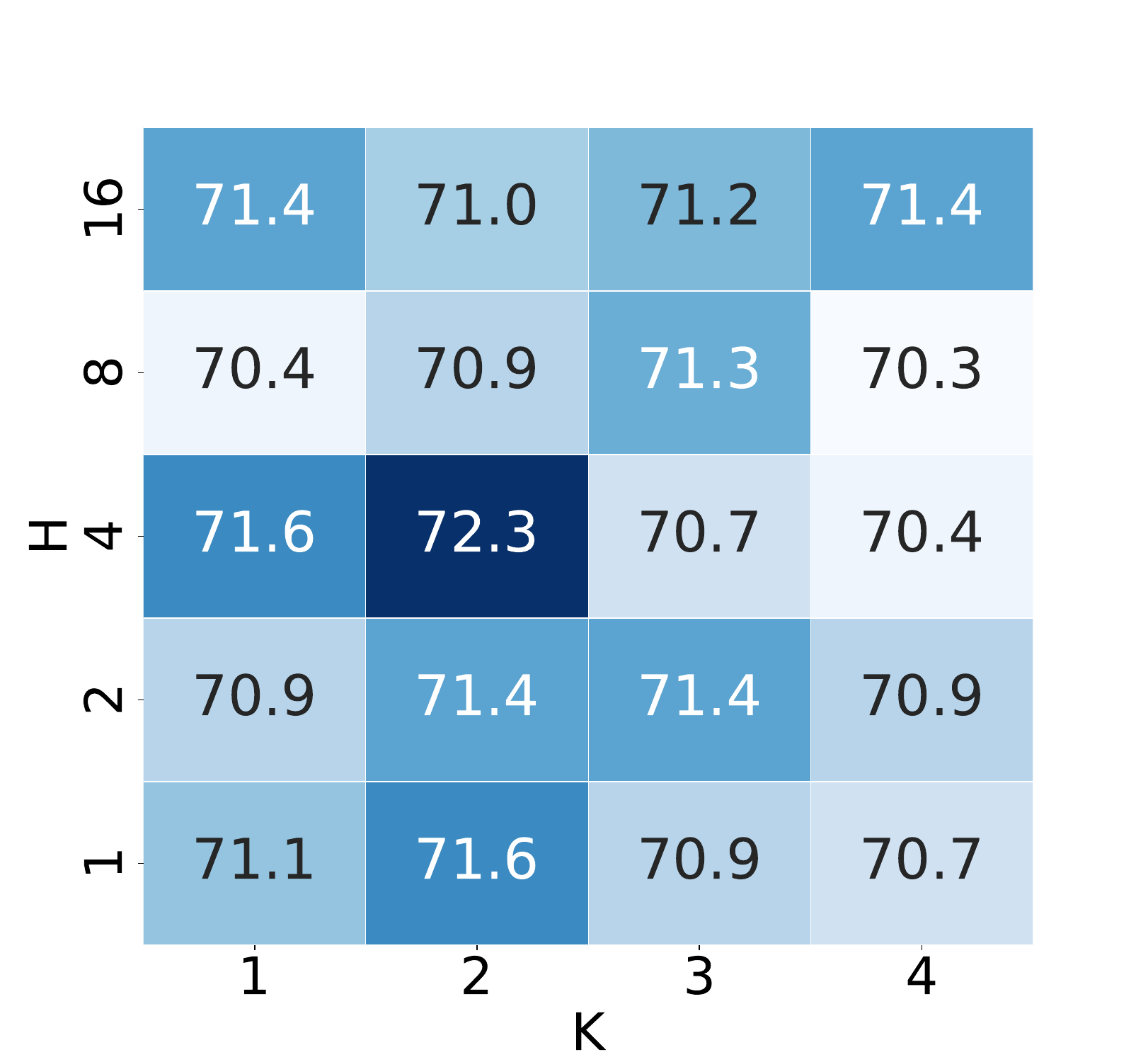}
    }
    \subfigure[Cars196 with PA loss]{
        \includegraphics[width=0.25\linewidth]{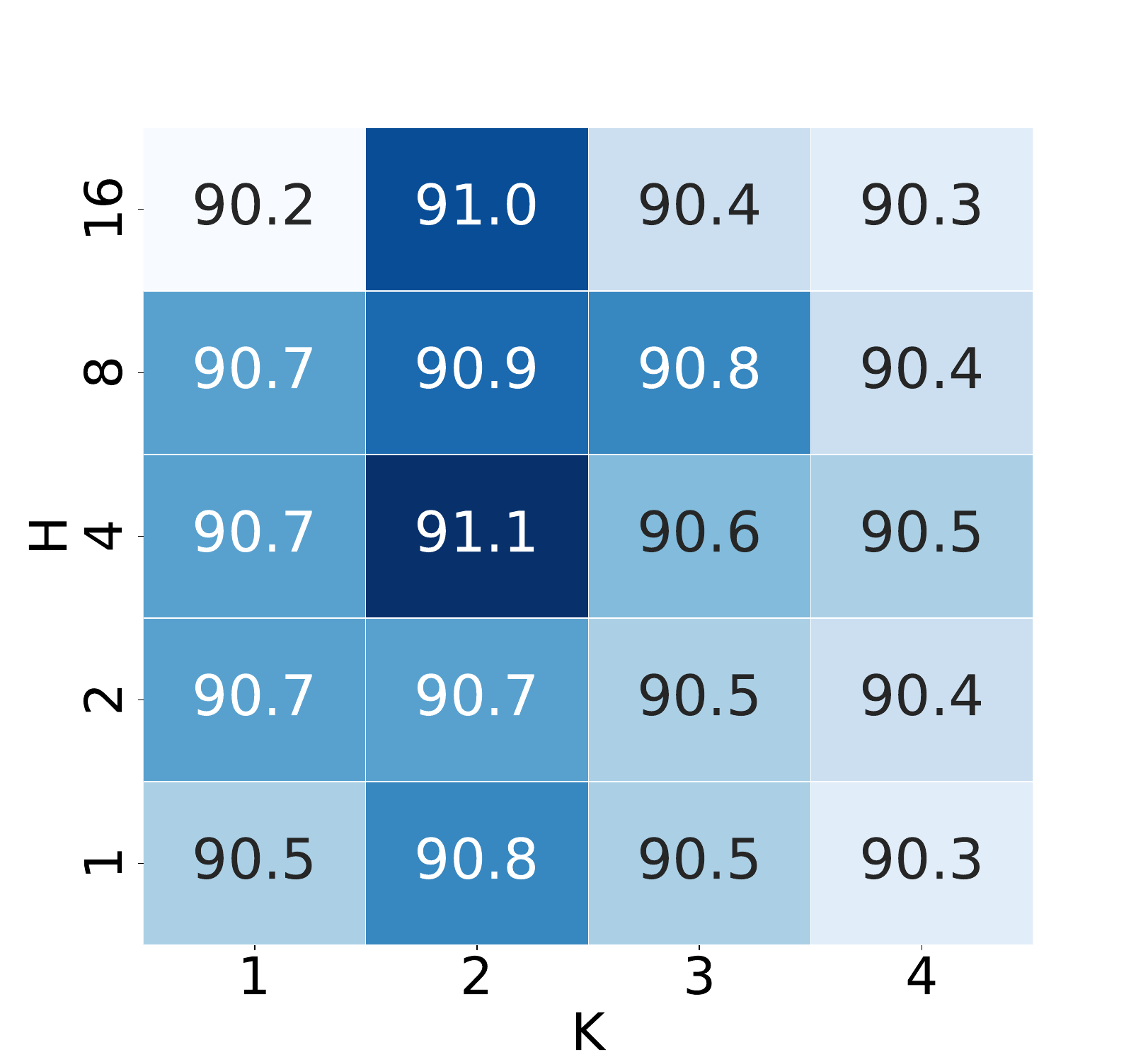}
    }
    \subfigure[InShop with PA loss]{
        \includegraphics[width=0.25\linewidth]{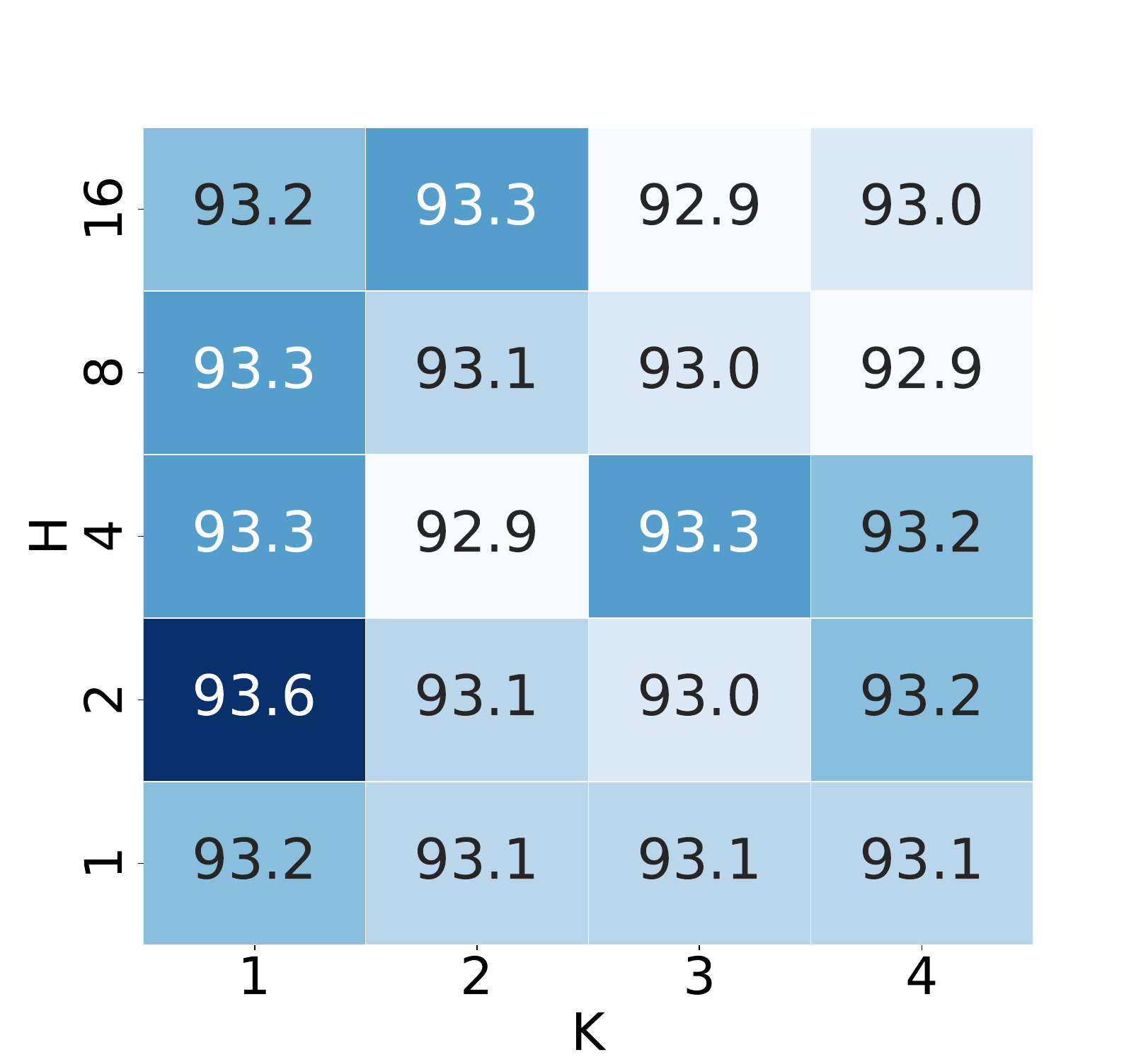}
    }
    \caption{Comparison of Recall@1 results versus the number of heads $H$ in the MMSA module and the iteration steps $K$ of graph message propagation with the setting of ResNet-50 combined with PA and our modified NP losses on the CUB-200-2011, Cars196, and InShop datasets.}
    \label{Fig_graph parameter}
\end{figure*}

To determine the optimal GNN configuration within our GCA-HNG framework, we analyzed the impact of varying structural parameters, focusing on the number of attention heads $H$ in the MMSA module and the iteration steps $K$ of graph message propagation. Given the significant disparities in data distributions of distinct datasets, selecting different hyper-parameters for network training is a common and effective strategy \cite{seidenschwarz2021learning,yang2023hse,elezi2022group}. While for datasets with similar distributions, we should maintain parameter consistency to enhance the model’s generalization across them. Based on the collective results in Fig. \ref{Fig_graph parameter}, we derived our configuration choices to balance model performance and structural generalization effectively. For the CUB-200-2011 and Cars196 datasets, which contain fewer classes but numerous samples per class and exhibit similar inter-class characteristics with substantial intra-class variation, we adopt a configuration with $K=2$ and $H=4$. This deeper graph structure enhances the model's ability to capture these complex patterns, which is crucial for datasets with closely related samples. In contrast, the InShop dataset includes many classes but fewer samples per class and tends to show lower inter-class similarity within the sampled batch. Therefore, we select a more streamlined configuration with $K=1$ and $H=2$ to efficiently handle this data distribution with higher discrimination and mitigate overfitting risks. We also apply this setup to the SOP dataset as they share similar class distributions, ensuring effective handling of datasets with numerous but sparsely distributed classes. These settings enable GCA-HNG to consistently achieve competitive performance across these diverse datasets.

\subsection{Analysis on the Robustness of GCA-HNG and Generator-based HNG Algorithms}
\label{experiment_bb}
\begin{table*}
\caption{Comparisons of image retrieval performance across the combinations of five network architectures with our modified N-pair loss on the CUB-200-2011 and Cars196 datasets.}
\centering
\tabcolsep=2mm
\begin{tabular}{lcccc|ccc}
\toprule
 \multirow{2}{*}{Method} & \multirow{2}{*}{Backbone} & \multicolumn{3}{c}{CUB-200-2011} & \multicolumn{3}{c}{Cars196} \\ \cmidrule{3-8}
  &  &  R@1 & RP & M@R & R@1 & RP & M@R  \\ \midrule
\midrule
Baseline         & G\_AVG   & 62.0$\pm$0.3 & 33.5$\pm$0.3 & 22.5$\pm$0.3 & 78.1$\pm$0.4 & 34.7$\pm$0.3 & 23.8$\pm$0.4 \\ 
Generator-based  & G\_AVG   & 63.5$\pm$0.1 & 33.5$\pm$0.2 & 22.4$\pm$0.1 & 82.9$\pm$0.1 & 35.1$\pm$0.1 & 24.6$\pm$0.1 \\ 
GCA-HNG          & G\_AVG   & 63.9$\pm$0.1 & 34.9$\pm$0.2 & 23.8$\pm$0.2 & 84.1$\pm$0.3 & 37.9$\pm$0.1 & 27.5$\pm$0.1 \\  \midrule
Baseline         & G\_MAX   & 63.4$\pm$0.1 & 34.3$\pm$0.2 & 23.4$\pm$0.1 & 78.7$\pm$0.2 & 34.7$\pm$0.3 & 23.8$\pm$0.2 \\ 
Generator-based  & G\_MAX   & 63.9$\pm$0.1 & 34.9$\pm$0.2 & 23.9$\pm$0.2 & 80.6$\pm$0.1 & 34.5$\pm$0.1 & 23.8$\pm$0.1 \\ 
GCA-HNG          & G\_MAX   & 64.3$\pm$0.1 & 35.0$\pm$0.3 & 24.0$\pm$0.2 & 83.8$\pm$0.1 & 37.2$\pm$0.3 & 26.8$\pm$0.2 \\  \midrule
Baseline         & R50\_AVG & 64.5$\pm$0.1 & 36.1$\pm$0.1 & 25.0$\pm$0.2 & 80.8$\pm$0.2 & 37.3$\pm$0.1 & 26.4$\pm$0.1 \\ 
Generator-based  & R50\_AVG & 65.4$\pm$0.1 & 35.8$\pm$0.1 & 24.6$\pm$0.1 & 83.1$\pm$0.1 & 38.3$\pm$0.2 & 27.8$\pm$0.1 \\ 
GCA-HNG          & R50\_AVG & 66.7$\pm$0.1 & 36.8$\pm$0.2 & 25.7$\pm$0.2 & 85.4$\pm$0.2 & 40.1$\pm$0.1 & 29.8$\pm$0.1 \\  \midrule
Baseline         & R50\_MAX & 66.5$\pm$0.1 & 36.5$\pm$0.3 & 25.5$\pm$0.3 & 84.8$\pm$0.1 & 39.5$\pm$0.1 & 29.3$\pm$0.1 \\ 
Generator-based  & R50\_MAX & 66.9$\pm$0.1 & 37.4$\pm$0.1 & 26.3$\pm$0.1 & 84.9$\pm$0.1 & 38.3$\pm$0.3 & 27.8$\pm$0.3 \\ 
GCA-HNG          & R50\_MAX & 70.3$\pm$0.3 & 39.4$\pm$0.2 & 28.4$\pm$0.3 & 88.2$\pm$0.3 & 42.3$\pm$0.4 & 32.5$\pm$0.4 \\  \midrule
Baseline         & DINO     & 76.7$\pm$0.2 & 43.6$\pm$0.2 & 33.0$\pm$0.3 & 87.6$\pm$0.1 & 39.7$\pm$0.2 & 29.7$\pm$0.1 \\ 
Generator-based  & DINO     & 77.1$\pm$0.1 & 43.9$\pm$0.2 & 33.2$\pm$0.2 & 86.9$\pm$0.1 & 38.9$\pm$0.1 & 28.7$\pm$0.1 \\ 
GCA-HNG          & DINO     & 78.5$\pm$0.1 & 44.4$\pm$0.2 & 33.9$\pm$0.1 & 90.0$\pm$0.1 & 43.6$\pm$0.1 & 34.1$\pm$0.1 \\
\bottomrule
\end{tabular}
\label{table_network}
\end{table*}

As deep learning evolves, a growing diversity of network architectures emerges. Given this variation, it is imperative for a HNG approach to demonstrate robust compatibility across these different backbones. In this section, we provide an in-depth analysis of the adaptability of our GCA-HNG framework, comparing it with existing Generator-based HNG algorithms across a range of backbone architectures.

In the context of HNG in deep metric learning, conventional methods typically incorporate a generator module that implements representation mapping to synthesize hard negatives \cite{zheng2019hardness,huang2020relationship,duan2019deep}. Such methods primarily focus on the representations derived from CNNs with average pooling, optimizing using pairwise loss. However, we observe that applying Generator-based HNG methods across varied network architectures and output layer designs poses significant challenges. To thoroughly investigate this aspect, we compared our GCA-HNG with conventional Generator-based HNG methods, which typically use a generator network to synthesize hard negatives from fused sample representations. In this comparison, we utilize the ``single coefficient interp.'' method as a representative example of sample fusion techniques prevalent in Generator-based HNG approaches, detailed in Table \ref{table_interpolation manner}. Subsequently, we evaluated performance using our modified NP loss combined with five distinct backbone and output layer configurations on the CUB-200-2011 and Cars196 datasets. These configurations expand beyond typical CNNs with average pooling used in Generator-based HNG methods, introducing both max pooling variants and a ViT architecture, including GoogLeNet with average pooling (G\_AVG), GoogLeNet with max pooling (G\_MAX), ResNet-50 with average pooling (R50\_AVG), ResNet-50 with max pooling (R50\_MAX), and DINO without any output processing (DINO). As illustrated in Table \ref{table_network}, the Generator-based method exhibits improvements in configurations with average pooling, whereas enhancements are negligible or absent in max pooling variants and DINO architectures. In contrast, GCA-HNG exhibits consistent robustness across all settings, highlighting its adaptability and efficacy. 

\begin{figure}
    \centering
    \subfigure[Comparison between GoogLeNet and DINO]{
        \includegraphics[width=0.97\linewidth]{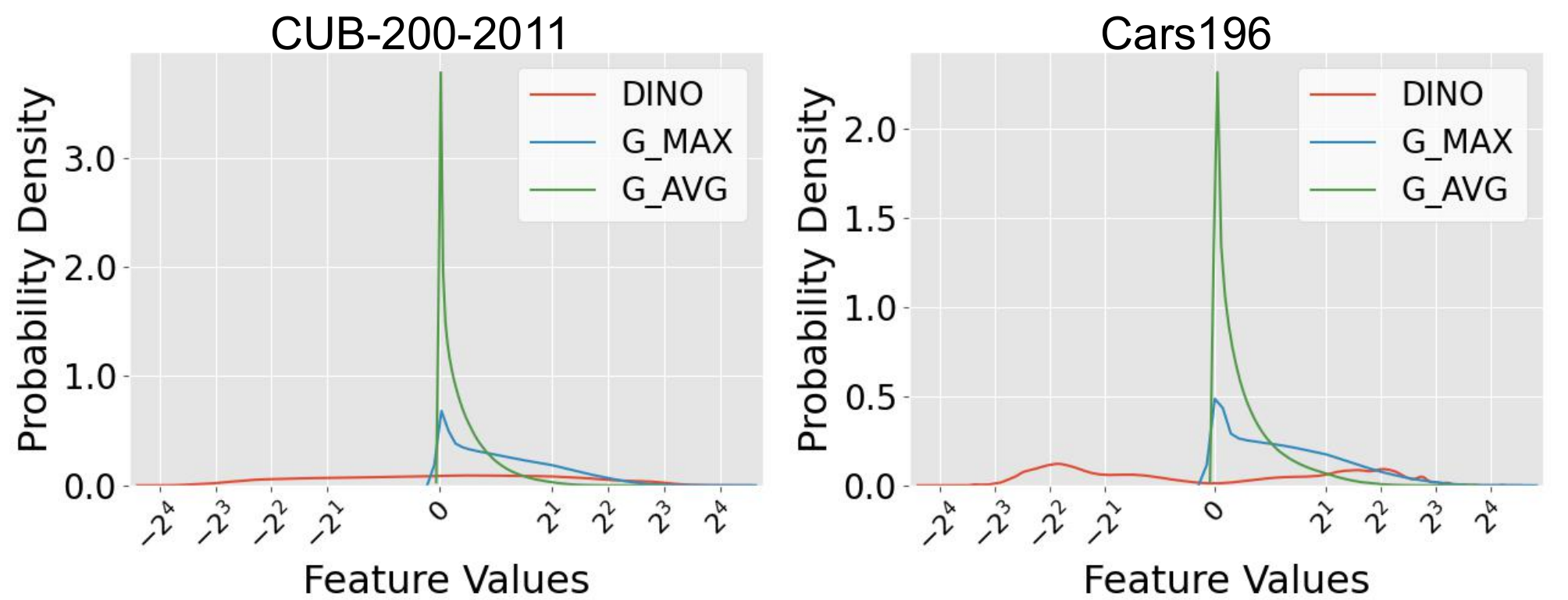}
    }
    \\
    \subfigure[Comparison between ResNet-50 and DINO]{
        \includegraphics[width=0.97\linewidth]{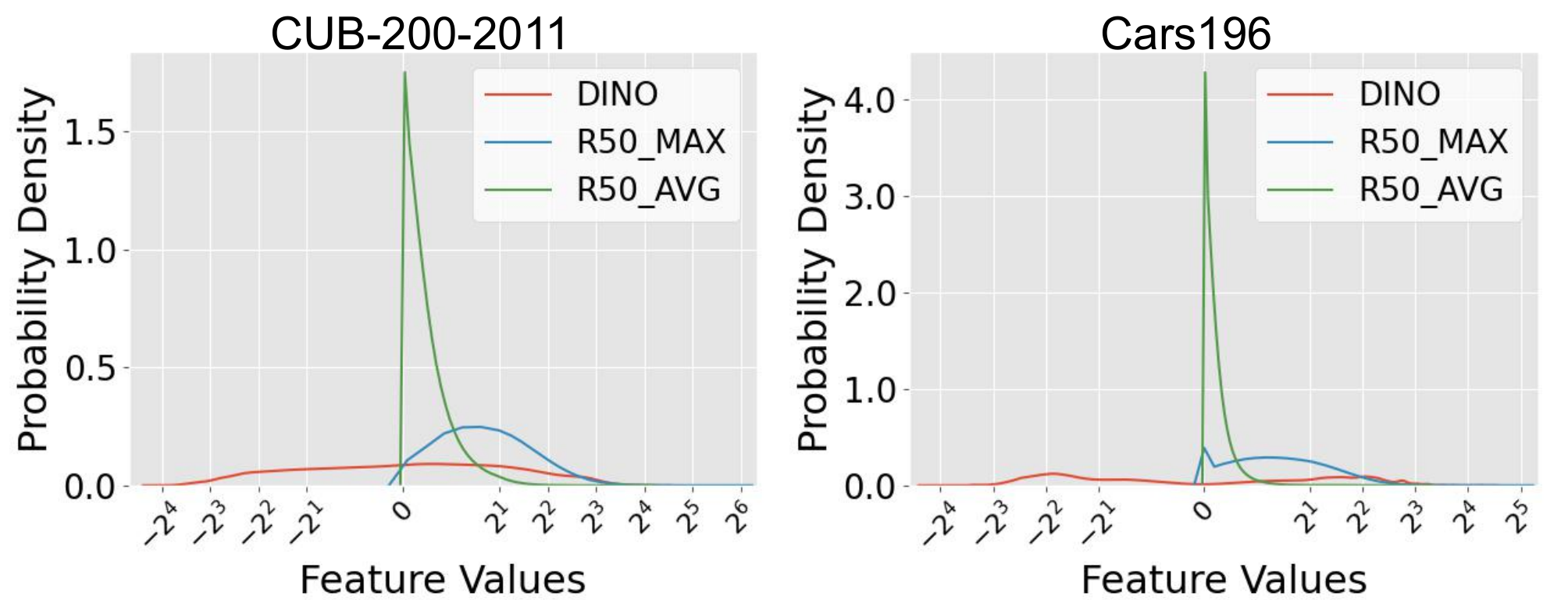}
    }
    \caption{Feature distribution analysis across five backbone architectures on the CUB-200-2011 and Cars196 datasets, explicitly comparing GoogLeNet (with max and average pooling) and DINO, as well as ResNet-50 (with max and average pooling) and DINO for clarity.}
    \label{Fig_feature distribution}
\end{figure}

To delve deeply into the performance disparities induced by different architectural backbones on the above two comparison methods, we analyzed feature distributions across the aforementioned five combinations of backbone networks and output layer designs, utilizing the testing sets of the CUB-200-2011 and Cars196 datasets. As depicted in Fig. \ref{Fig_feature distribution}, within an identical CNN backbone, max pooling leads to larger variances in feature distribution compared to average pooling. Furthermore, the DINO architecture demonstrates substantially larger feature variance than the other four CNN configurations. In conjunction with the retrieval performance presented in Table \ref{table_network}, a notable observation emerges: an increase in feature variance tends to correlate with a reduced improvement for Generator-based HNG methods, whereas the GCA-HNG approach remains unaffected, indicative of its robustness against diverse feature variances. This divergence in performance stems from the inherent properties of the Generator-based methods, which typically utilize MSE loss for feature alignment optimization within the generator module. This optimization strategy tends to be more effective with values close to the standard normal distribution but struggles across broader ranges marked by high uncertainty. As a result, this limitation restricts the applicability of these Generator-based HNG methods to various complex and diverse backbone structures.

In summary, our approach shifts the focus from traditional generator networks to a learnable interpolation vector, thereby circumventing the need for an additional generator for feature alignment optimization. This endeavor seeks to demonstrate the wide applicability of GCA-HNG and its potential as a flexible solution in the evolving realm of metric learning.

\section{Limitation}
While the GCA-HNG framework effectively utilizes global sample correlations to generate informative negatives, it primarily captures these correlations within the confines of a mini-batch. Due to hardware constraints, this limitation means it fails to consider the broader context of correlations across all classes simultaneously. Consequently, the framework may struggle to synthesize hard negatives between the categories most requiring differentiation, as they are often not sampled together within a batch. This issue becomes particularly significant in datasets with a large number of categories. Future work could explore methods to incorporate sample correlations on a larger scale to facilitate HNG. By considering correlations across all classes, we may better perceive nearby similar classes and synthesize informative negatives to help push class boundaries apart, thus learning more discriminative metrics.

\section{Conclusion}
In this paper, we propose a Globally Correlation-Aware Hard Negative Generation (GCA-HNG) framework, which designs a GCL module to learn sample correlations from a global perspective and introduces a CACAI module to generate hardness-adaptive and diverse negatives. The proposed GCA-HNG constructs a structured graph to model sample correlations and employs a GNN combined with the iterative graph message propagation mechanism to learn sample correlations globally. Leveraging the learned global sample correlations, we further propose a channel-adaptive interpolation manner to integrate an anchor with multiple negatives from a specific class for HNG. These synthetic hard negatives are crucial for facilitating the model to learn more discriminative metrics. We have validated the adaptability and efficacy of GCA-HNG by integrating numerous typical backbone architectures and metric losses, testing it across four widely used image retrieval benchmark datasets, and conducting comprehensive ablation studies on its core components. \\
\textbf{Code availability.} The codes and trained models are publicly available on GitHub: \url{https://github.com/PWenJay/GCA-HNG}. \\
\textbf{Data availability.} The data that support the finding of this study are openly available at the following URLs: \url{https://data.caltech.edu/records/65de6-vp158}, \url{https://ai.stanford.edu/~jkrause/cars/car_dataset.html}, \url{https://cvgl.stanford.edu/projects/lifted_struct/}, \url{https://mmlab.ie.cuhk.edu.hk/projects/DeepFashion.html}.

\begin{acknowledgements}
The research is partially supported by National Key Research and Development Program of China (2023YFC3502900), National Natural Science Foundation of China (No.62176093, 61673182, 62206060), Key Realm Research and Development Program of Guangzhou (No.202206030001), Guangdong-Hong Kong-Macao Joint Innovation Project (No.2023A0505030016).
\end{acknowledgements}

\bibliographystyle{spmpsci}      
\bibliography{reference} 

\begin{thebibliography}{10}
\providecommand{\url}[1]{{#1}}
\providecommand{\urlprefix}{URL }
\expandafter\ifx\csname urlstyle\endcsname\relax
  \providecommand{\doi}[1]{DOI~\discretionary{}{}{}#1}\else
  \providecommand{\doi}{DOI~\discretionary{}{}{}\begingroup \urlstyle{rm}\Url}\fi

\bibitem{aziere2019ensemble}
Aziere, N., Todorovic, S.: Ensemble deep manifold similarity learning using hard proxies.
\newblock In: CVPR, pp. 7299--7307 (2019)

\bibitem{bucher2016hard}
Bucher, M., Herbin, S., Jurie, F.: Hard negative mining for metric learning based zero-shot classification.
\newblock In: ECCV, pp. 524--531. Springer (2016)

\bibitem{caron2021emerging}
Caron, M., Touvron, H., Misra, I., J{\'e}gou, H., Mairal, J., Bojanowski, P., Joulin, A.: Emerging properties in self-supervised vision transformers.
\newblock In: ICCV, pp. 9650--9660 (2021)

\bibitem{chen2022knowledge}
Chen, T., Lin, L., Chen, R., Hui, X., Wu, H.: Knowledge-guided multi-label few-shot learning for general image recognition.
\newblock IEEE Transactions on Pattern Analysis and Machine Intelligence \textbf{44}(3), 1371--1384 (2022)

\bibitem{chen2024heterogeneous}
Chen, T., Pu, T., Liu, L., Shi, Y., Yang, Z., Lin, L.: Heterogeneous semantic transfer for multi-label recognition with partial labels.
\newblock International Journal of Computer Vision \textbf{132}, 6091–6106 (2024)

\bibitem{chen2022cross}
Chen, T., Pu, T., Wu, H., Xie, Y., Liu, L., Lin, L.: Cross-domain facial expression recognition: A unified evaluation benchmark and adversarial graph learning.
\newblock IEEE Transactions on Pattern Analysis and Machine Intelligence \textbf{44}(12), 9887--9903 (2022)

\bibitem{chen2024dynamic}
Chen, T., Wang, W., Pu, T., Qin, J., Yang, Z., Liu, J., Lin, L.: Dynamic correlation learning and regularization for multi-label confidence calibration.
\newblock IEEE Transactions on Image Processing \textbf{33}, 4811--4823 (2024)

\bibitem{chen2019multi}
Chen, Z.M., Wei, X.S., Wang, P., Guo, Y.: Multi-label image recognition with graph convolutional networks.
\newblock In: CVPR, pp. 5177--5186 (2019)

\bibitem{dai2025on}
Dai, G., Zhang, Y., Ke, Q., Guo, Q., Huang, S.: One-dm: One-shot diffusion mimicker for handwritten text generation.
\newblock In: ECCV, pp. 410--427 (2025)

\bibitem{dai2023disentangling}
Dai, G., Zhang, Y., Wang, Q., Du, Q., Yu, Z., Liu, Z., Huang, S.: Disentangling writer and character styles for handwriting generation.
\newblock In: CVPR, pp. 5977--5986 (2023)

\bibitem{dosovitskiy2020image}
Dosovitskiy, A., Beyer, L., Kolesnikov, A., Weissenborn, D., Zhai, X., Unterthiner, T., Dehghani, M., Minderer, M., Heigold, G., Gelly, S., et~al.: An image is worth 16x16 words: Transformers for image recognition at scale.
\newblock arXiv preprint arXiv:2010.11929  (2020)

\bibitem{duan2019deep}
Duan, Y., Lu, J., Zheng, W., Zhou, J.: Deep adversarial metric learning.
\newblock IEEE Transactions on Image Processing \textbf{29}, 2037--2051 (2019)

\bibitem{elezi2022group}
Elezi, I., Seidenschwarz, J., Wagner, L., Vascon, S., Torcinovich, A., Pelillo, M., Leal-Taixe, L.: The group loss++: A deeper look into group loss for deep metric learning.
\newblock IEEE Transactions on Pattern Analysis and Machine Intelligence \textbf{45}(2), 2505--2518 (2022)

\bibitem{elezi2020group}
Elezi, I., Vascon, S., Torcinovich, A., Pelillo, M., Leal-Taix{\'e}, L.: The group loss for deep metric learning.
\newblock In: ECCV, pp. 277--294. Springer (2020)

\bibitem{ermolov2022hyperbolic}
Ermolov, A., Mirvakhabova, L., Khrulkov, V., Sebe, N., Oseledets, I.: Hyperbolic vision transformers: Combining improvements in metric learning.
\newblock In: CVPR, pp. 7409--7419 (2022)

\bibitem{gajic2021fast}
Gaji{\'c}, B., Amato, A., Gatta, C.: Fast hard negative mining for deep metric learning.
\newblock Pattern Recognition \textbf{112}, 107,795 (2021)

\bibitem{gu2020symmetrical}
Gu, G., Ko, B.: Symmetrical synthesis for deep metric learning.
\newblock In: AAAI, pp. 10,853--10,860 (2020)

\bibitem{gu2021proxy}
Gu, G., Ko, B., Kim, H.G.: Proxy synthesis: Learning with synthetic classes for deep metric learning.
\newblock In: AAAI, pp. 1460--1468 (2021)

\bibitem{hadsell2006dimensionality}
Hadsell, R., Chopra, S., LeCun, Y.: Dimensionality reduction by learning an invariant mapping.
\newblock In: CVPR, pp. 1735--1742 (2006)

\bibitem{he2016deep}
He, K., Zhang, X., Ren, S., Sun, J.: Deep residual learning for image recognition.
\newblock In: CVPR, pp. 770--778 (2016)

\bibitem{huang2022agtgan}
Huang, H., Yang, D., Dai, G., Han, Z., Wang, Y., Lam, K.M., Yang, F., Huang, S., Liu, Y., He, M.: Agtgan: Unpaired image translation for photographic ancient character generation.
\newblock In: ACM MM, pp. 5456--5467 (2022)

\bibitem{huang2020relationship}
Huang, J., Feng, Y., Zhou, M., Qiang, B.: Relationship-aware hard negative generation in deep metric learning.
\newblock In: KSEM, pp. 388--400. Springer (2020)

\bibitem{husain2021actnet}
Husain, S.S., Ong, E.J., Bober, M.: Actnet: end-to-end learning of feature activations and multi-stream aggregation for effective instance image retrieval.
\newblock International Journal of Computer Vision \textbf{129}, 1432--1450 (2021)

\bibitem{ioffe2015batch}
Ioffe, S., Szegedy, C.: Batch normalization: Accelerating deep network training by reducing internal covariate shift.
\newblock In: ICML, pp. 448--456. pmlr (2015)

\bibitem{jin2018unsupervised}
Jin, S., RoyChowdhury, A., Jiang, H., Singh, A., Prasad, A., Chakraborty, D., Learned-Miller, E.: Unsupervised hard example mining from videos for improved object detection.
\newblock In: ECCV, pp. 307--324 (2018)

\bibitem{kearnes2016molecular}
Kearnes, S., McCloskey, K., Berndl, M., Pande, V., Riley, P.: Molecular graph convolutions: moving beyond fingerprints.
\newblock J. Comput. Aided Mol. Des \textbf{30}, 595--608 (2016)

\bibitem{kim2023hier}
Kim, S., Jeong, B., Kwak, S.: Hier: Metric learning beyond class labels via hierarchical regularization.
\newblock In: CVPR, pp. 19,903--19,912 (2023)

\bibitem{kim2020proxy}
Kim, S., Kim, D., Cho, M., Kwak, S.: Proxy anchor loss for deep metric learning.
\newblock In: CVPR, pp. 3238--3247 (2020)

\bibitem{kipfsemi}
Kipf, T.N., Welling, M.: Semi-supervised classification with graph convolutional networks.
\newblock In: ICLR (2017)

\bibitem{ko2020embedding}
Ko, B., Gu, G.: Embedding expansion: Augmentation in embedding space for deep metric learning.
\newblock In: CVPR, pp. 7255--7264 (2020)

\bibitem{krause20133d}
Krause, J., Stark, M., Deng, J., Fei-Fei, L.: 3d object representations for fine-grained categorization.
\newblock In: ICCVW, pp. 554--561 (2013)

\bibitem{li2022self}
Li, D., Wang, Z., Wang, J., Zhang, X., Ding, E., Wang, J., Zhang, Z.: Self-guided hard negative generation for unsupervised person re-identification.
\newblock In: IJCAI (2022)

\bibitem{li2020weakly}
Li, Z., Tang, J., Zhang, L., Yang, J.: Weakly-supervised semantic guided hashing for social image retrieval.
\newblock International Journal of Computer Vision \textbf{128}, 2265--2278 (2020)

\bibitem{liao2022graph}
Liao, S., Shao, L.: Graph sampling based deep metric learning for generalizable person re-identification.
\newblock In: CVPR, pp. 7359--7368 (2022)

\bibitem{lim2022hypergraph}
Lim, J., Yun, S., Park, S., Choi, J.Y.: Hypergraph-induced semantic tuplet loss for deep metric learning.
\newblock In: CVPR, pp. 212--222 (2022)

\bibitem{lin2024contrastive}
Lin, Z., Li, J., Dai, G., Chen, T., Huang, S., Lin, J.: Contrastive representation enhancement and learning for handwritten mathematical expression recognition.
\newblock Pattern Recognition Letters \textbf{186}, 14--20 (2024)

\bibitem{liu2020learning}
Liu, H., Wang, R., Shan, S., Chen, X.: Learning multifunctional binary codes for personalized image retrieval.
\newblock International Journal of Computer Vision \textbf{128}(8-9), 2223--2242 (2020)

\bibitem{liu2016deepfashion}
Liu, Z., Luo, P., Qiu, S., Wang, X., Tang, X.: Deepfashion: Powering robust clothes recognition and retrieval with rich annotations.
\newblock In: CVPR, pp. 1096--1104 (2016)

\bibitem{loshchilov2018decoupled}
Loshchilov, I., Hutter, F.: Decoupled weight decay regularization.
\newblock In: ICLR (2018)

\bibitem{lu2017discriminative}
Lu, J., Hu, J., Tan, Y.P.: Discriminative deep metric learning for face and kinship verification.
\newblock IEEE Transactions on Image Processing \textbf{26}(9), 4269--4282 (2017)

\bibitem{movshovitz2017no}
Movshovitz-Attias, Y., Toshev, A., Leung, T.K., Ioffe, S., Singh, S.: No fuss distance metric learning using proxies.
\newblock In: ICCV, pp. 360--368 (2017)

\bibitem{musgrave2020metric}
Musgrave, K., Belongie, S., Lim, S.N.: A metric learning reality check.
\newblock In: ECCV, pp. 681--699. Springer (2020)

\bibitem{oh2016deep}
Oh~Song, H., Xiang, Y., Jegelka, S., Savarese, S.: Deep metric learning via lifted structured feature embedding.
\newblock In: CVPR, pp. 4004--4012 (2016)

\bibitem{qian2019softtriple}
Qian, Q., Shang, L., Sun, B., Hu, J., Li, H., Jin, R.: Softtriple loss: Deep metric learning without triplet sampling.
\newblock In: ICCV, pp. 6450--6458 (2019)

\bibitem{rao2023hierarchical}
Rao, H., Leung, C., Miao, C.: Hierarchical skeleton meta-prototype contrastive learning with hard skeleton mining for unsupervised person re-identification.
\newblock International Journal of Computer Vision pp. 1--23 (2023)

\bibitem{roth2022non}
Roth, K., Vinyals, O., Akata, Z.: Non-isotropy regularization for proxy-based deep metric learning.
\newblock In: CVPR, pp. 7420--7430 (2022)

\bibitem{russakovsky2015imagenet}
Russakovsky, O., Deng, J., Su, H., Krause, J., Satheesh, S., Ma, S., Huang, Z., Karpathy, A., Khosla, A., Bernstein, M., et~al.: Imagenet large scale visual recognition challenge.
\newblock International Journal of Computer Vision \textbf{115}(3), 211--252 (2015)

\bibitem{schroff2015facenet}
Schroff, F., Kalenichenko, D., Philbin, J.: Facenet: A unified embedding for face recognition and clustering.
\newblock In: CVPR, pp. 815--823 (2015)

\bibitem{seidenschwarz2021learning}
Seidenschwarz, J.D., Elezi, I., Leal-Taix{\'e}, L.: Learning intra-batch connections for deep metric learning.
\newblock In: ICML, pp. 9410--9421. PMLR (2021)

\bibitem{shrivastava2016training}
Shrivastava, A., Gupta, A., Girshick, R.: Training region-based object detectors with online hard example mining.
\newblock In: CVPR, pp. 761--769 (2016)

\bibitem{simo2015discriminative}
Simo-Serra, E., Trulls, E., Ferraz, L., Kokkinos, I., Fua, P., Moreno-Noguer, F.: Discriminative learning of deep convolutional feature point descriptors.
\newblock In: ICCV, pp. 118--126 (2015)

\bibitem{smirnov2018hard}
Smirnov, E., Melnikov, A., Oleinik, A., Ivanova, E., Kalinovskiy, I., Luckyanets, E.: Hard example mining with auxiliary embeddings.
\newblock In: CVPRW, pp. 37--46 (2018)

\bibitem{sohn2016improved}
Sohn, K.: Improved deep metric learning with multi-class n-pair loss objective.
\newblock In: NeurIPS, pp. 1857--1865 (2016)

\bibitem{suh2019stochastic}
Suh, Y., Han, B., Kim, W., Lee, K.M.: Stochastic class-based hard example mining for deep metric learning.
\newblock In: CVPR, pp. 7251--7259 (2019)

\bibitem{szegedy2015going}
Szegedy, C., Liu, W., Jia, Y., Sermanet, P., Reed, S., Anguelov, D., Erhan, D., Vanhoucke, V., Rabinovich, A.: Going deeper with convolutions.
\newblock In: CVPR, pp. 1--9 (2015)

\bibitem{tan2022cross}
Tan, Z., Liu, A., Wan, J., Liu, H., Lei, Z., Guo, G., Li, S.Z.: Cross-batch hard example mining with pseudo large batch for id vs. spot face recognition.
\newblock IEEE Transactions on Image Processing \textbf{31}, 3224--3235 (2022)

\bibitem{teh2020proxynca++}
Teh, E.W., DeVries, T., Taylor, G.W.: Proxynca++: Revisiting and revitalizing proxy neighborhood component analysis.
\newblock In: ECCV, pp. 448--464. Springer (2020)

\bibitem{touvron2021training}
Touvron, H., Cord, M., Douze, M., Massa, F., Sablayrolles, A., J{\'e}gou, H.: Training data-efficient image transformers \& distillation through attention.
\newblock In: ICML, pp. 10,347--10,357. PMLR (2021)

\bibitem{vaswani2017attention}
Vaswani, A., Shazeer, N., Parmar, N., Uszkoreit, J., Jones, L., Gomez, A.N., Kaiser, {\L}., Polosukhin, I.: Attention is all you need.
\newblock NeurIPS \textbf{30} (2017)

\bibitem{velivckovicgraph}
Veli{\v{c}}kovi{\'c}, P., Cucurull, G., Casanova, A., Romero, A., Li{\`o}, P., Bengio, Y.: Graph attention networks.
\newblock In: ICLR (2018)

\bibitem{venkataramanan2022takes}
Venkataramanan, S., Psomas, B., Kijak, E., Amsaleg, L., Karantzalos, K., Avrithis, Y.: It takes two to tango: Mixup for deep metric learning.
\newblock In: ICLR, pp. 1--21 (2022)

\bibitem{Wang_2023_CVPR}
Wang, C., Zheng, W., Li, J., Zhou, J., Lu, J.: Deep factorized metric learning.
\newblock In: CVPR, pp. 7672--7682 (2023)

\bibitem{wang2017deep}
Wang, J., Zhou, F., Wen, S., Liu, X., Lin, Y.: Deep metric learning with angular loss.
\newblock In: ICCV, pp. 2593--2601 (2017)

\bibitem{wang2019multi}
Wang, X., Han, X., Huang, W., Dong, D., Scott, M.R.: Multi-similarity loss with general pair weighting for deep metric learning.
\newblock In: CVPR, pp. 5022--5030 (2019)

\bibitem{wang2020multi}
Wang, Y., He, D., Li, F., Long, X., Zhou, Z., Ma, J., Wen, S.: Multi-label classification with label graph superimposing.
\newblock In: AAAI, vol.~34, pp. 12,265--12,272 (2020)

\bibitem{weinberger2009distance}
Weinberger, K.Q., Saul, L.K.: Distance metric learning for large margin nearest neighbor classification.
\newblock Journal of Machine Learning Research \textbf{10}(2) (2009)

\bibitem{welinder2010caltech}
Welinder, P., Branson, S., Mita, T., Wah, C., Schroff, F., Belongie, S., Perona, P.: Caltech-ucsd birds 200.
\newblock California Institute of Technology  (2010)

\bibitem{xuan2020improved}
Xuan, H., Stylianou, A., Pless, R.: Improved embeddings with easy positive triplet mining.
\newblock In: WACV, pp. 2474--2482 (2020)

\bibitem{yang2023hse}
Yang, B., Sun, H., Li, F.W., Chen, Z., Cai, J., Song, C.: Hse: Hybrid species embedding for deep metric learning.
\newblock In: ICCV, pp. 11,047--11,057 (2023)

\bibitem{yang2019learning}
Yang, L., Zhan, X., Chen, D., Yan, J., Loy, C.C., Lin, D.: Learning to cluster faces on an affinity graph.
\newblock In: CVPR, pp. 2298--2306 (2019)

\bibitem{yang2022hierarchical}
Yang, Z., Bastan, M., Zhu, X., Gray, D., Samaras, D.: Hierarchical proxy-based loss for deep metric learning.
\newblock In: WACV, pp. 1859--1868 (2022)

\bibitem{yu2021fine}
Yu, Q., Song, J., Song, Y.Z., Xiang, T., Hospedales, T.M.: Fine-grained instance-level sketch-based image retrieval.
\newblock International Journal of Computer Vision \textbf{129}, 484--500 (2021)

\bibitem{zeng2022keyword}
Zeng, Y., Wang, Y., Liao, D., Li, G., Huang, W., Xu, J., Cao, D., Man, H.: Keyword-based diverse image retrieval with variational multiple instance graph.
\newblock IEEE Transactions on Neural Networks and Learning Systems \textbf{34}(12), 10,528--10,537 (2022)

\bibitem{zhai2018classification}
Zhai, A., Wu, H.Y.: Classification is a strong baseline for deep metric learning.
\newblock arXiv preprint arXiv:1811.12649  (2018)

\bibitem{zhang2022attributable}
Zhang, B., Zheng, W., Zhou, J., Lu, J.: Attributable visual similarity learning.
\newblock In: CVPR, pp. 7532--7541 (2022)

\bibitem{zhang2023denoising}
Zhang, C., Luo, L., Gu, B.: Denoising multi-similarity formulation: a self-paced curriculum-driven approach for robust metric learning.
\newblock In: AAAI, vol.~37, pp. 11,183--11,191 (2023)

\bibitem{zhao2018adversarial}
Zhao, Y., Jin, Z., Qi, G.j., Lu, H., Hua, X.s.: An adversarial approach to hard triplet generation.
\newblock In: ECCV, pp. 501--517 (2018)

\bibitem{zheng2019hardness}
Zheng, W., Chen, Z., Lu, J., Zhou, J.: Hardness-aware deep metric learning.
\newblock In: CVPR, pp. 72--81 (2019)

\bibitem{zheng2021hardness}
Zheng, W., Lu, J., Zhou, J.: Hardness-aware deep metric learning.
\newblock IEEE Transactions on Pattern Analysis and Machine Intelligence \textbf{43}(9), 3214--3228 (2021)

\bibitem{zhu2022construct}
Zhu, C., Hu, Z., Dong, H., He, G., Yu, Z., Zhang, S.: Construct informative triplet with two-stage hard-sample generation.
\newblock Neurocomputing \textbf{498}, 59--74 (2022)

\bibitem{zhu2023attribute}
Zhu, J., Liu, L., Zhan, Y., Zhu, X., Zeng, H., Tao, D.: Attribute-image person re-identification via modal-consistent metric learning.
\newblock International Journal of Computer Vision \textbf{131}(11), 2959--2976 (2023)

\bibitem{zhu2021visual}
Zhu, S., Yang, T., Chen, C.: Visual explanation for deep metric learning.
\newblock IEEE Transactions on Image Processing \textbf{30}, 7593--7607 (2021)

\bibitem{zhu2019distance}
Zhu, X., Jing, X.Y., Zhang, F., Zhang, X., You, X., Cui, X.: Distance learning by mining hard and easy negative samples for person re-identification.
\newblock Pattern Recognition \textbf{95}, 211--222 (2019)

\bibitem{zhu2020fewer}
Zhu, Y., Yang, M., Deng, C., Liu, W.: Fewer is more: a deep graph metric learning perspective using fewer proxies.
\newblock In: NeurIPS, pp. 17,792--17,803 (2020)

\end{thebibliography}

\end{sloppypar}
\end{document}